\pdfoutput=1
\documentclass[11pt]{article}

\usepackage[]{emnlp2021}
\usepackage{times}
\usepackage{latexsym}

\usepackage[T1]{fontenc}
\usepackage[utf8]{inputenc}

\usepackage{graphicx}
\usepackage{amsmath,bm}
\usepackage{amsfonts}
\usepackage{multirow}
\usepackage{booktabs}
\usepackage{microtype}
\usepackage{adjustbox}
\usepackage{tikz}
\usepackage{pgf}
\usepackage{pgfplots}
\usepackage{tikz-qtree}
\usetikzlibrary{arrows,decorations.pathmorphing,backgrounds,positioning,fit,petri,shapes.misc, arrows.meta,shapes.geometric,decorations.markings,calc,shadows.blur,decorations.pathreplacing,quotes,matrix,shapes.symbols}
\definecolor{fontgray}{RGB}{44, 62, 80}
\definecolor{myred}{RGB}{235, 47, 6} %rgb()
\definecolor{summertime}{RGB}{245, 205, 121}
\definecolor{darkgrass}{RGB}{0, 148, 50}
\definecolor{myblue}{RGB}{0, 168, 255}
\definecolor{mygray}{RGB}{158, 158, 158}
\definecolor{puffin}{RGB}{250, 152, 58}
\definecolor{lowpurple}{RGB}{210, 180, 222}
\definecolor{lowblue}{RGB}{102,178,255}
\definecolor{lowred}{RGB}{245, 183, 177}
\definecolor{customblue}{RGB}{14, 128, 136}
\definecolor{customred}{RGB}{184, 84, 80}

\newcounter{Lcount}
\newcommand{\squishenum}{
 \begin{list}{\arabic{Lcount}. }
  { \usecounter{Lcount}
   \setlength{\itemsep}{0pt}
   \setlength{\parsep}{0pt}
   \setlength{\topsep}{0pt}
   \setlength{\partopsep}{0pt}
   \setlength{\leftmargin}{2em}
   \setlength{\labelwidth}{1.5em}
   \setlength{\labelsep}{0.5em} } }
 
 \newcommand{\squishletter}{
  \begin{list}{\alph{Lcount}. }
   { \usecounter{Lcount}
    \setlength{\itemsep}{0pt}
    \setlength{\parsep}{0pt}
    \setlength{\topsep}{0pt}
    \setlength{\partopsep}{0pt}
    \setlength{\leftmargin}{2em}
    \setlength{\labelwidth}{1.5em}
    \setlength{\labelsep}{0.5em} } }
  
  \newcommand{\squishlist}{
   \begin{list}{$\bullet$}
    { \usecounter{Lcount}
     \setlength{\itemsep}{0pt}
     \setlength{\parsep}{0pt}
     \setlength{\topsep}{0pt}
     \setlength{\partopsep}{0pt}
     \setlength{\leftmargin}{2em}
     \setlength{\labelwidth}{1.5em}
     \setlength{\labelsep}{0.5em} } }

   \newcommand{\squishend}{
  \end{list} }
  
\title{To be Closer: Learning to Link up Aspects with Opinions}% for Aspect-based Sentiment Analysis}
\author{Yuxiang Zhou$^{1*}$, Lejian Liao$^{1}$, Yang Gao$^{1\dag}$, Zhanming Jie$^{2,3}$ \and Wei Lu$^{2}$ \\
        \textsuperscript{1}School of Computer Science and Technology, Beijing Institute of Technology \\ \textsuperscript{2}StatNLP Research Group, Singapore University of Technology and Design\\
        \textsuperscript{3}ByteDance AI Lab\\
        \texttt{yxzhou@bit.edu.cn,} \texttt{liaolj@bit.edu.cn,} \texttt{gyang@bit.edu.cn}\\
        \texttt{allan@bytedance.com,} \texttt{luwei@sutd.edu.sg}}
\begin{document}
\maketitle
\renewcommand{\thefootnote}{\fnsymbol{footnote}}
\footnotetext[1]{Work done when visiting SUTD.}
\footnotetext[2]{Corresponding author.}
\footnotetext{Accepted as a long paper in the main conference of EMNLP 2021 (Conference on Empirical Methods in Natural Language Processing).\\}
\renewcommand{\thefootnote}{\arabic{footnote}}
% \footnotetext[2]{Corresponding author.}

%%%abs
\begin{abstract}
Dependency parse trees are helpful for discovering the opinion words in aspect-based sentiment analysis (ABSA)~\cite{huang2019syntax}.
However, the trees obtained from off-the-shelf dependency parsers are static, and could be sub-optimal in ABSA. 
This is because the syntactic trees are not designed for capturing the interactions between opinion words and aspect words.
In this work, we aim to shorten the distance between aspects and corresponding opinion words by learning an aspect-centric tree structure.
The aspect and opinion words are expected to be closer along such tree structure compared to the standard dependency parse tree.
The learning process allows the tree structure to adaptively correlate the aspect and opinion words, enabling us to better identify the polarity in the ABSA task.
We conduct experiments on five aspect-based sentiment datasets, and the proposed model significantly outperforms recent strong baselines.
Furthermore, our thorough analysis demonstrates the average distance between aspect and opinion words are shortened by at least 19\% on the standard SemEval Restaurant14~\citep{Pontiki2014SemEval2014T4} dataset.
% Specifically, our model achieves new state-of-the-art results on the standard SemEval Restaurant15 and Restaurant16 datasets. 
% We further conduct a thorough analysis to justify the effectiveness of the aspect-centric latent trees. 
% Our findings demonstrate that it is essential to model the entire latent trees for promoting the interactions between the aspects and opinions\footnote{We release our code at **URL**.}. 
% \textcolor{red}{(Allan: better to add conclusive findings)}\footnote{Our code is available at \url{https://github.com/zyxnlp/ACLT}.}
\end{abstract}

%%%intro
\section{Introduction}
Aspect-based sentiment analysis (ABSA)~\citep{pang2008opinion,liu2012sentiment} aims at determining the sentiment polarity expressed towards a particular target in a sentence. 
For example, in the sentence~{\em``The \textbf{battery life} of this laptop is very long, but the \textbf{price} is too high''}, the sentiment expressed towards the aspect term {\em ``battery life''} is positive, whereas the sentiment towards the aspect term~{\em ``price''} is negative. 
% ABSA is extensively studied both in academic communities and industry since it makes it possible to provide fine-grained analysis of the users’ opinion towards the specific aspect that is present in the text. 
% The main challenge of ABSA is to effectively capture the interaction between aspect and context words. 
% For example, identifying the opinion words  {\em``long"} and {\em``high"} which express the polarity of the aspect terms {\em ``battery life"} and {\em ``price"}, respectively. 
\begin{figure}[t!]
	\centering
	\adjustbox{max width=1.0\linewidth}{
		\begin{tikzpicture}[node distance=2.0mm and 2.0mm, >=Stealth, 
			wordnode/.style={draw=none, minimum height=5mm, inner sep=0pt},
			chainLine/.style={line width=1pt,-, color=fontgray},
			entbox/.style={draw=black, rounded corners, fill=red!20, dashed}
			]
			%		\node [word](w1) [] {\footnotesize Ah};
			%		\node [word, right=of w1](w2) [] {\footnotesize ,};
			%		\node [word, right=of w2](w3) [] {\footnotesize today};
			%		\node [word, right=of w3](w4) [] {\footnotesize is};
			\matrix(sent1) [matrix of nodes, row sep=35mm, column sep=5mm, nodes in empty cells, execute at empty cell=\node{\strut};]
			{
				\textcolor{customblue}{\textbf{\textit{\Large  Loving }}} & [1mm]\textit{\Large  the } &[1mm]\textcolor{customred}{\textbf{\textit{\Large harry}}}  & [1mm] \textcolor{customred}{\textbf{\textit{\Large potter}}}   &  [1mm] \textit{\Large  movie }& [1mm] \textit{\Large marathon...}  \\
				\textcolor{customblue}{\textbf{\textit{\Large  Loving }}} & [1mm]\textit{\Large  the } &[1mm]\textcolor{customred}{\textbf{\textit{\Large harry}}}  & [1mm] \textcolor{customred}{\textbf{\textit{\Large potter}}}   &  [1mm] \textit{\Large  movie }& [1mm] \textit{\Large marathon...}   \\
			};
			
%			\draw[chainLine, ->, color=fontgray, line width=1.5pt] (sent1-1-1) to [out=60,in=120, looseness=0.8] node[above, yshift=-1mm, color=black]{} (sent1-1-7);
			\draw[chainLine, ->, customblue] (sent1-1-1) to [out=60,in=120, looseness=0.6] node[above, yshift=-1mm, color=black]{} (sent1-1-6);
			
			\draw[chainLine, <-] (sent1-1-2) to [out=60,in=120, looseness=0.6] node[above, yshift=-1mm, color=black]{} (sent1-1-6);
			\draw[chainLine, <-,customblue] (sent1-1-3) to [out=60,in=120, looseness=1] node[above, yshift=-1mm, color=black]{} (sent1-1-4);
			%			\draw[chainLine, ->] (sent1-1-4) to [out=60,in=120, looseness=1] node[above, yshift=-1mm, color=black]{} (sent1-1-8);
			\draw [chainLine, <-, customblue] (sent1-1-4) to [out=60,in=120, looseness=1] node[above, yshift=-1mm, xshift=-3mm, color=black]{} (sent1-1-5);
			\draw [chainLine, <-, customblue] (sent1-1-5) to [out=60,in=120, looseness=1] node[above, yshift=-1mm, xshift=-3mm, color=black]{} (sent1-1-6);

%			\draw[chainLine, ->, color=fontgray, line width=1.5pt] (sent1-2-1) to [out=60,in=120, looseness=0.8] node[above, yshift=-1mm, color=black]{} (sent1-2-7);
			\draw[chainLine, ->] (sent1-2-1) to [out=60,in=120, looseness=0.6] node[above, yshift=-1mm, color=black]{} (sent1-2-6);
			\draw[chainLine, <-, customblue] (sent1-2-1) to [out=60,in=120, looseness=0.8] node[above, yshift=-1mm, color=black]{} (sent1-2-3);
			\draw[chainLine, <-] (sent1-2-2) to [out=60,in=120, looseness=1] node[above, yshift=-1mm, color=black]{} (sent1-2-4);
			\draw[chainLine, <-, customblue] (sent1-2-3) to [out=60,in=120, looseness=1] node[above, yshift=-1mm, color=black]{} (sent1-2-4);
			\draw[chainLine, ->] (sent1-2-4) to [out=60,in=120, looseness=1] node[above, yshift=-1mm, color=black]{} (sent1-2-5);
			%			\draw[chainLine, ->] (sent1-2-3) to [out=60,in=120, looseness=1] node[above, yshift=-1mm, color=black]{} (sent1-2-8);

			\node[wordnode, above= of sent1-1-1, yshift=10mm] (root1) {\Large \texttt{root}};
			\draw[chainLine, -] (root1) to  (sent1-1-1);
			
			\node[wordnode, above= of sent1-2-4, yshift=7mm] (root2) {\Large \texttt{root}};
			\draw[chainLine, -] (root2) to  (sent1-2-4);

			\node[wordnode, below= of sent1,yshift=3mm] (b) {\Large (b) Learned aspect-centric Tree (Ours)};
			\node[wordnode, above= of b, yshift=34mm] (a) {\Large (a) Dependency parse tree from spaCy.};

			\node[wordnode, above= of sent1, yshift=16mm] (da) {\large  {Dist(\textit{Loving}, \textit{harry}) = 4,~~  Dist(\textit{Loving}, \textit{potter}) = 3}};
			\node[wordnode, below= of da, yshift=-37mm] (db)  {\large  {Dist(\textit{Loving}, \textit{harry}) = 1,~~  Dist(\textit{Loving}, \textit{potter}) = 2}};
		\end{tikzpicture} 
	}
% 		\vspace*{-6mm}%%cr version
%	\caption{An example from the Twitter dataset to illustrate the difference between (a) a dependency parse tree  and (b) an aspect-centric tree. Red words indicate the aspect words of the sentence.}
	\caption{An example with different tree representations in the Twitter dataset.
		``Dist'' returns the number of hops between two words in the tree.
		Words marked in red and blue are aspect and opinion, respectively. }
% 		\vspace*{-6mm}%%cr version
	\label{fig:intro}
\end{figure}
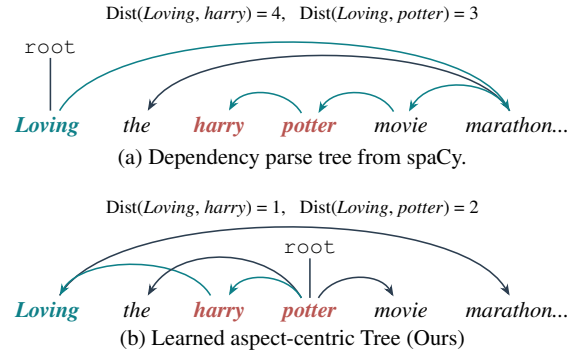
Early research efforts~\citep{wang2016attention,chen2017recurrent,liu2017attention,li2019learning,xu2020aspect} focus on using an attention mechanism~\citep{bahdanau2014neural} to model interactions between aspect and context words. 
However, such attention-based models may suffer from overly focusing on the frequent words that express sentiment polarity while ignoring low-frequency ones~\citep{tang2019progressive,sun-lu-2020-understanding}. 
% It has been shown in prior efforts that the syntactic structures of sentences can facilitate identification of sentiment features related to aspect words~\citep{zhang2019aspect,sun2019aspect,huang2019syntax}. 
Recent efforts show that the syntactic structures of sentences can facilitate the identification of sentiment features related to aspect words~\citep{zhang2019aspect,sun2019aspect,huang2019syntax}.
Nonetheless, these methods unfortunately suffer from two shortcomings. 
First, the trees obtained from off-the-shelf dependency parsers are static, and thus cannot adaptively model the complex relationship between multiple aspects and opinion words. 
Second, an inaccurate parse tree could lead to error propagation downstream in  the pipeline.
Several research groups have explored the above issues with a more refined parse tree.
For example, ~\citet{chen2020inducing} constructed task-specific structures by developing a gate mechanism to dynamically combine the parse tree information and a stochastic graph sampled from the HardKuma distribution~\citep{bastings2019interpretable}.
On the other hand, ~\citet{wang2020relational} greedily reshaped the dependency parse tree by using manual rules to obtain the aspect-related syntactic structures.

% Recently, several research efforts~\citep{huang2019syntax,wang2020relational,chen2020inducing} show that dependency structure of sentence facilitate modeling interaction between aspect and context words.

% inconsistent may choose other word?
\textcolor{black}{Despite being able to effectively alleviate the tree representation problem, existing methods still depend on external parse trees, leading to one potential problem.
The dependency parse trees are not designed for the purpose of ABSA but to express syntactic relations.
Specifically, the aspect term is usually a noun or a noun phrase, while the root of the dependency tree is often a verb or an adverb. 
According to statistics, for almost 90\%\footnote{The detail statistic can be found in Appendix~\ref{app:A}.} of the sentences, the roots of their dependency trees are not aspect words.
% almost 90\%\footnote{The detail statistic can be found in Appendix A.1.} of sentences whose root of the dependency tree is not aspect words.
Such a \textit{root inconsistency} issue may prevent the model from effectively capturing the relationships between opinion words and aspect words.}
% {\color{red}
% Specifically, the aspect term is usually a noun or noun phrase while the root of the dependency tree is often a verb or an adverb. 
% % If the dependency tree of the sentence is not rooted in either aspect word in the sentence, we define it as a root inconsistency issue.
% Given a sentence $\bm{s} $ and the corresponding aspect term $\bm{a}$, if the dependency tree of $\bm{s}$ is not rooted in either aspect word $w_i \in \bm{a}$, we define it as a \textbf{\textit{root inconsistency}} issue.}
% {\color{red}Such a \textit{root inconsistency} issue may prevent the model from effectively capturing the relationships between opinion words and aspect words.}
For example, Figure~\ref{fig:intro}(a) shows the dependency tree obtained by the toolkit spaCy\footnote{\url{https://spacy.io/api/dependencyparser}}. 
The root is the gerund verb ``\textit{Loving}'' while the aspect term is the noun phrase ``\textit{harry potter}''.
% According to statistics, this inconsistency phenomenon exists for almost 90\%\footnote{The detail statistic can be found in Appendix A.1.} of sentences in each dataset we used.
% Second, the dependency tree may push the opinion words away from the aspect words.
The distance between the aspect words ``\textit{harry}'' and  ``\textit{potter}'' and the critical opinion word ``\textit{Loving}'' under a dependency tree are four hops and three hops, respectively.
However, their relative distances in the sequential order are two and three, respectively.
Intuitively, closer distance enables us to identify the polarity in the ABSA task better.
% {\color{red}While prior efforts on using latent (or explicit) trees for the ABSA task exist, one of the major contributions of our work is that we identified the root inconsistency issue.}
Figure~\ref{fig:intro}(b) shows an aspect-centric tree where the tree is rooted by the aspect words.
The distances between aspect and opinion words are one hop and two hops, which is closer than the distance in the standard dependency parse tree. 
In this paper, we propose a model that learns {\em Aspect-Centric Latent Trees} (we name it as the ACLT model) which are specifically tailored for the ABSA task.
% {\color{red}The model tackles the aforementioned problems, which treats the dependency tree as a latent variable and provides supervision to refine the root to aspect word, thus shorten the distance between the aspect and corresponding opinion words.}
\textcolor{black}{We assume that inducing tree structures whose roots are within aspect term enables the model to correlate the aspect and opinion words better. 
We built our model based on the structure attention mechanism~\citep{kim2017structured,liu2018learning} and a variant of the Matrix-Tree Theorem (MTT)~\citep{tutte1984graph,koo2007structured}.
% Additionally, we impose explicit constraints on aspect words to improve the probability of them becoming the root when inducing a tree.
% Additionally, we proposed to impose a constraint to make the aspect words more likely to be the root of the tree structure induced by MTT. 
Additionally, we proposed to impose a soft constraint to encourage the aspect words to serve as the root of the tree structure induced by MTT.
As a result, the search space of inferring the root is reduced during the training process. Our code is available at \url{https://github.com/zyxnlp/ACLT}.}
% In other words, we reduce the search space of inferring root for MTT in the training process.
% Specifically, we compute each aspect word's probability to become the root which restraint

% {\color{blue}As it is common for the aspect term to have multiple words in our task, we compute each aspect word's probability to become the root rather than computing the probability of a single aspect word.
% Thus, the roots of the trees induced by MTT could only be the aspect words rather than the others.

% In other words, we lower the boundary of inferencing root for MTT in the training process.
% We name our proposed model ACLT, short for Aspect-Centric Latent Tree model.
% Moreover, we compute the marginal probabilities of multi-root to dynamically build the latent tree based on multiple aspect words, allowing the model to capture complex interactions among words within aspect term.  

% We further develop a multi-root constraint strategy, which enables our model to dynamically build the latent tree based on multiple aspect words, allowing the model to capture complex interactions. We name our proposed model ACLT, short for Aspect-Centric Latent Tree model.

% Experiments show that our ACLT model outperforms the existing approaches on the ABSA benchmark datasets. Specifically, our model surpasses the current state-of-the-art models on the standard SemEval Restaurant15~\citep{pontiki2015semeval} and Restaurant16~\citep{pontiki2016semeval} datasets by 4.21\% and 2.02\% in terms of F1 score, respectively\footnote{We release our code at **URL**.}. 
Our contributions are summarized as follows:
% root constrain is able to transfer the root of latent tree to aspect words, thus generate a aspect-centric latent tree.
% improve the probability of them becoming the root when inducing a matrix tree
\begin{itemize}

% \item We construct a sentence-level graph for inference in an end-to-end fashion without relying on an external parser or manual rules. 
% Leveraging word-level signals, our model is able to dynamically construct an aspect-centric latent structure to establish an effective connection between aspect and opinion words. 

% is there a stronger or more specific description we can use there? accurate? reliable?

% \item Treating the aspect words as the root, we learn a aspect-centric tree representation to shorten the distance between the aspect and corresponding opinion words. 
% Such representation allows us to effectively capture the interactions between aspect and opinion words.
% \item We tailor a novel latent-variable model, which is able to help us learn the structural bias to facilitate the interactions between target and opinion words in ABSA task. 
\item We propose to use {\em Aspect-Centric Latent Trees} (ACLT) which are specifically tailored for the ABSA task to link up aspects with opinion words in an end-to-end fashion.
\item Our ACLT model is able to learn an aspect-centric latent tree with a root refinement strategy to better correlate the aspect and opinion words than the standard parse tree.
\item Experiments show that our model outperforms the existing approaches, and also yields new state-of-the-art results on four ABSA benchmark datasets.
Quantitative and qualitative experiments further justify the effectiveness of the learned aspect-centric trees.
% with state-of-the-art models in various settings. 
The analysis demonstrates that our ACLT is capable of shortening the average distances between aspect and opinion words by at least 19\% on the standard SemEval Restaurant14 dataset.
To the best of our knowledge, we are the first to link up aspects with opinions through the specifically designed latent tree that imposes root constraints. %%cr version

% discovering a more appropriate tree structure by adjusting the aspects to be the roots, and thus shortening the distance between aspects and opinions. 
% Empirical results show that the multi-root constraint strategy enables our model effectively captures the complex interactions among words within aspect term. %% task-specific rewrite，specific conclusion
% thorough analysis demonstrates our model shortens the average distances between aspect and opinion words by at least 19\% in the standard SemEval Restaurant14~\citep{pontiki-etal-2014-semeval}

\end{itemize}
%Experiment results showing that the proposed model performs competitively against the state-of-the-art model without relying on an external parser. 

%%%method
\section{Model}
\begin{figure}[t!]
   \centering
   {\includegraphics[width=0.5\textwidth]{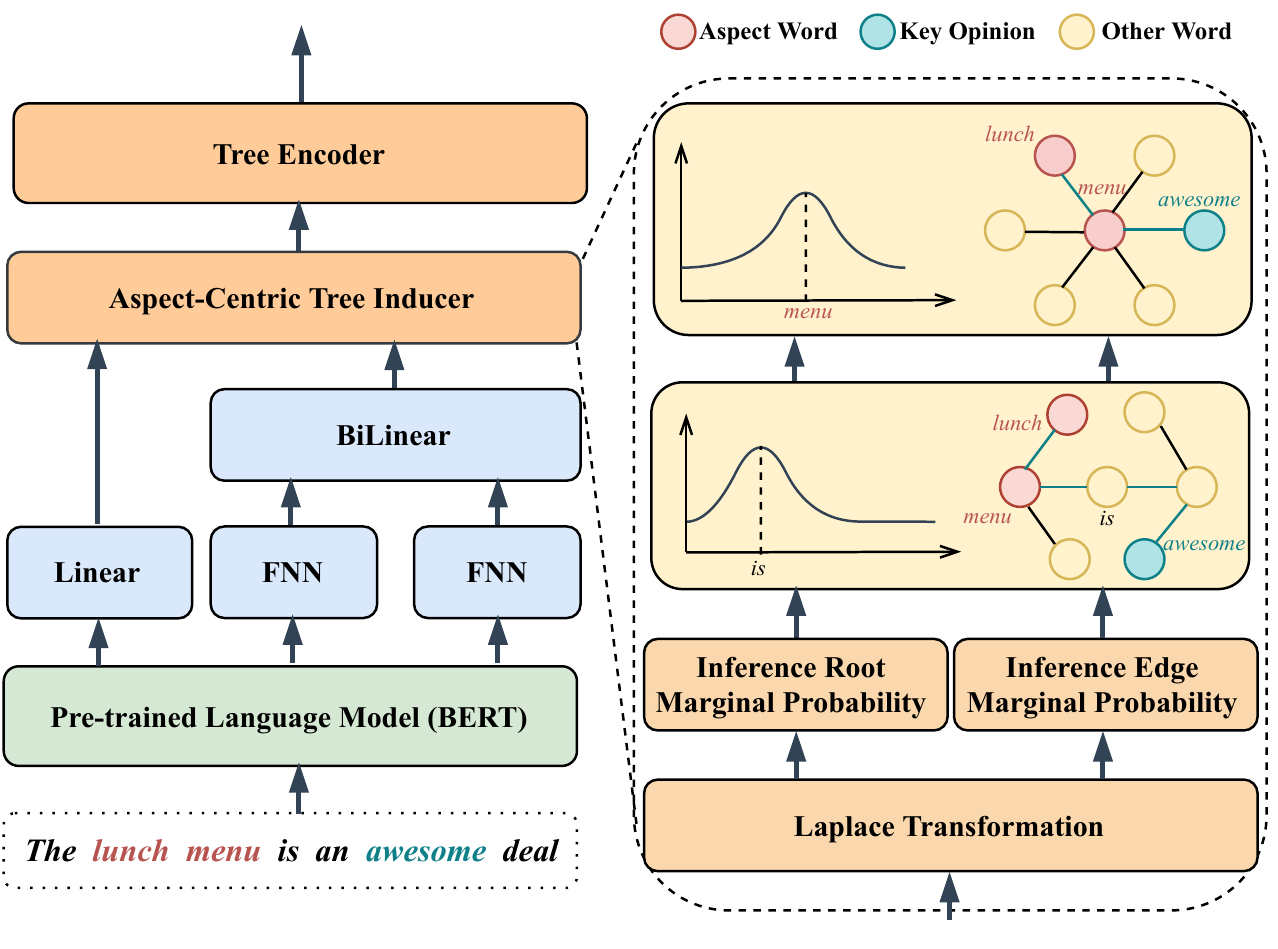}\label{vis(architure)}}
%   \vspace{-5mm}%%cr version
   \caption{ACLT architecture.}
%   \vspace{-5mm}%%cr version
%   \caption{Overview of our proposed ACLT model. It consists of three components. A sentence encoder is applied to get the contextualized representations of sentences. The tree inducer, shown in detail to the right bottom of the figure. The root refinement, and the aspect-centric tree encoder.}
   \label{fig:model}
\end{figure}

In this section, we present the proposed Aspect-Centric Latent Tree (ACLT) model (Figure~\ref{fig:model}) for the ABSA task.
% Given an input sentence $\bm{s} =\{w_1,...,w_n\}$ and a aspect term as $\bm{a} =\{w_i,...,w_j \}$ ($1 \leq i \leq j \leq n$), where $n$ is the length of the sentence, the goal of ABSA is to predict sentiment polarity $y \in \{ \textit{positive}, \textit{neutral},\textit{negative}\}$ over the given aspect. 
% Figure~\ref{fig:model} illustrates the overall architecture of our ACLT model.
We first obtain the contextualized representations from the sentence encoder.
% , which can be a pre-trained language model BERT in our experiment.
% The sentence encoder first encodes each word of the input sentence and outputs contextual representations. 
Next, we use a tree inducer to produce the distribution over all the possible latent trees. 
% our tree inducer produce the aspect-centric tree structure given the contextualized representations.
The underlying tree inducer is a latent-variable model which treats tree structures as the latent variable.
% The tree inducer is then applied to induce an aspect-centric latent structure based on the representation of each word (node) and the aspect signal. 
% Representations of nodes are updated based on information propagation on the latent structure. 
Once we have the distribution over the latent trees, we adopt the root refinement procedure to obtain aspect-centric latent trees. 
Then, we can encode the probabilistic latent trees with a graph or tree encoder.
% In the tree encoder, the aspect-centric latent trees are encoded to the contextual representations to generate the aspect-specific structured representations. 
Finally, we use the structured representation from the tree encoder for sentiment classification. 
\subsection{Sentence Encoder} 
% \paragraph{Sentence Encoder}
Given a sentence $\bm{s} =[w_1,...,w_n]$ and the corresponding aspect term  $\bm{a} =[w_i,...,w_j]$ ($1 \leq i \leq j \leq n$), we adopt the pre-trained language model BERT~\citep{devlin2019bert} to obtain the contextualized representation for each word. 
We concatenate the words in the sentence and explicitly present the aspect term in the input representation:  $\mathbf{x} = ( \texttt{[CLS]} ~ w_1,...,w_n ~ \texttt{[SEP]} ~ w_i,...,w_j ~ \texttt{[SEP]} )$. 
The contextualized representation $\bm{H}$ can be obtained via $\text{BERT}(\mathbf{x})$, 
%% 我印象中input 是  句子 然后 [SEP] 然后 aspect 对吧 得重写一下。感觉不清楚
% \begin{equation}
% \begin{split}
%     \mathbf{w} & = w_1,...,w_n ~ \textsc{[SEP]} ~ w_i,...,w_j\\
%     \mathbf{x} & = \text{BERT}(\mathbf{x})\text{,} 
% \end{split}
% \end{equation}
% each word $w_i$ is fed to the sentence encoder, which outputs the contextualized representations of each word in $w_i$. 
% The sentence encoder can be a pre-train model such as BERT~\citep{devlin2019bert} or a traditional context encoder (e.g., GRU~\citep{cho2014learning}). Here we use BERT as an example:
% \begin{equation}
% 	\bm{H}= \textbf{BERT}(\bm{X})\text{,} 
% \end{equation}
where $\bm{H}=[\bm{h}_{1},...,\bm{h}_n]$, $\bm{h}_i \in \bm{H}$ represents the contextualized representation of the $i$-th token. 

\subsection{Aspect-Centric Tree Inducer}
% Unlike existing models that use manual rules~\citep{wang2020relational} or the combination of the external parser and latent structure~\citep{chen2020inducing} to construct an aspect-specific tree, our model treats the tree as a latent variable and induces it in an end-to-end fashion. 
While prior efforts~\citep{wang2020relational,chen2020inducing} on learning latent (or explicit) trees for the ABSA task exist, one of the major contributions of our work is that we link up aspects and opinion words by addressing the root inconsistency issue.
% The tree induction module is built based on the structured attention method~\citep{kim2017structured,liu2018learning}. 
Inspired by recent work~\citep{liu2018learning,nan2020reasoning}, we use a variant of Kirchhoff’s Matrix-Tree Theorem~\citep{tutte1984graph,koo2007structured} to induce the latent dependency structure.

Given the contextualized representation $\bm{h} \in \mathbb R^{d} $ of each node (token) in the sentence, where $d$ is the dimension of the node representations. We first calculate pair-wise unnormalized edge scores $\bm{e}_{ij}$ between the $i$-th and the $j$-th node with the node representation $\bm{h}_i$ and $\bm{h}_j$ by way of a two feed-forward neural network (FNN) and a bilinear function:
\begin{equation}
	\bm{e}_{ij}=\big( \tanh(\bm{W}_{p}\bm{h}_i))^T\bm{W}_{b}(\tanh(\bm{W}_{c}\bm{h}_j) \big)\text{,}
\end{equation}
where $\bm{W}_{p} \in \mathbb R^{d \times d}$ and $\bm{W}_{c} \in \mathbb R^{d \times d}$ are weights for two feedforward neural networks, tanh is applied as the activation function. $\bm{W}_{b} \in \mathbb R^{d \times d}$ is the weight for the bilinear transformation. $\bm{e}_{ij} \in \mathbb R^{d \times d} $ can be viewed as a weighted adjacency matrix for a graph $G$ with $n$ nodes where each node corresponds to a word in the sentence. 

Next, we calculate the root score $\bm{r}_{i}$, representing the unnormalized probability of the $i$-th node to be selected as the root of the structure:
\begin{equation}
	\bm{r}_{i}=\bm{W}_{r}\bm{h}_{i}\text{,}
\end{equation}
where $\bm{W}_{r} \in \mathbb R^{1 \times d}$ is the weight for the linear transformation. Following~\citet{koo2007structured}, we calculate the marginal probability of the dependency edge of the latent structure:
\begin{align}
& \boldsymbol{A}_{i j}=\left\{\begin{array}{ll}
0 & \text { if } i=j \\
\exp \left({\boldsymbol{e}}_{i j}\right) & \text { otherwise }
\end{array}\right. \\
& \boldsymbol{L}_{i j}=\left\{\begin{array}{ll}
\sum_{i^{\prime}=1}^{n} \boldsymbol{A}_{i^{\prime} j} & \text { if } i=j \\
-\boldsymbol{A}_{i j} & \text { otherwise }
\end{array}\right.\\
& \bar{\boldsymbol{L}}_{i j}=\left\{\begin{array}{ll}
\boldsymbol{L}_{i j}+\exp \left({\boldsymbol{r}}_{i}\right) &  \text{ if } i=j \\
\boldsymbol{L}_{i j} & \text { otherwise, }
\end{array}\right. \label{eq:multi}
\end{align}
where we first assign non-negative weights $\boldsymbol{A} \in \mathbb R^{n \times n}$ to the edges,  $\boldsymbol{A}_{i j}$ is the weight of the edge between the $i$-th and the $j$-th node. 
Then, we build the Laplacian matrix $\boldsymbol L \in \mathbb R^{n \times n}$ for graph $G$ and its variant $\bar{\boldsymbol{L}}$ which takes the root node into consideration for further computation~\citep{koo2007structured}. 
% To tackle this problem, we compute the marginal probabilities of a multi-root tree in the tree inducer module by Equation (\ref{eq:multi}) 
% Rather than computing the root probabilities for all words~\citep{liu2018learning,nan2020reasoning}, we compute the root probabilities for aspect word $w_i \in \bm{a}$, (Equation \ref{eq:multi}). 
% Thus, it is easier for our model to learn the space of all possible roots during training.
% Thus, we can lower the boundary of inferencing root of MTT in training process.
% In contrast to~\citep{liu2018learning,nan2020reasoning}, we compute the marginal probabilities only based on aspect word $w_i \in \bm{a}$. 
We use $\boldsymbol{P}_{i j}$ to denote the marginal probability of the dependency edge between the $i$-th and the $j$-th node,  and $\boldsymbol{P}_{i}^{r}$ is defined as the marginal probability of the $i$-th word headed by the root of the tree. 
Then, $\boldsymbol{P}_{i j}$ and $\boldsymbol{P}_{i}^{r}$ can be derived:
% \vspace{-2mm}%%cr version
\begin{equation}
\begin{aligned}
\boldsymbol{P}_{i j}=&\left(1-\delta_{1, j}\right) \boldsymbol{A}_{i j}\left[\bar{\boldsymbol{L}}^{-1}\right]_{j j} \\
&-\left(1-\delta_{i, 1}\right) \boldsymbol{A}_{i j}\left[\bar{\boldsymbol{L}}^{-1}\right]_{j i}\label{equ:adj}
\end{aligned}
% \vspace{-2mm}%%cr version
\end{equation}
\begin{equation}
\boldsymbol{P}_{i}^{r}=\exp \left({\boldsymbol{r}}_{i}\right)\left[\bar{\boldsymbol{L}}^{-1}\right]_{i 1}\text{,}
\label{equ:rootprob}
\end{equation}
where $\delta$ is the Kronecker delta. Here, $\boldsymbol{P} \in \mathbb R^{n \times n}$ can be interpreted as a weighted adjacency matrix of the word-level graph. We refer the interested reader to~\citet{koo2007structured} for more details. 

% If we need to highlight the importance, subsected it. But we only have text describe? Should we add a statistic to describe the number?
% Nevertheless, in the ABSA task, sentences in which the aspect term contains multiple words are common and generally challenging to settle. The reason for this is that we are not able to model the relation between the critical aspect word and context adaptively. To tackle this problem, we compute the marginal probabilities of a multi-root tree in the tree inducer module by Equation (\ref{eq:six}) rather than computing the marginal probabilities of a single-root tree~\citep{liu2018learning,nan2020reasoning}. 

% In contrast, to~\citet{nan2020reasoning}, who compute the marginal probabilities of a single-root tree, our tree has multiple roots since in our task, the aspect term may contains multiple words.%% need to describe the multi-root marginal probability

% \paragraph{Loss for Aspect Root}%% need improvement
\paragraph{Root Refinement}%% need improvement
% As previous works~\citep{liu2018learning,guo2020learning,nan2020reasoning} have shown, the MTT gives us some random tree structures if we do not provide proper guidance during training. %% todo , guidance?
% As previous works~\citep{liu2018learning,guo2020learning,nan2020reasoning} have shown, the MTT would produce arbitrary trees without any structure supervision.
Despite the successful application of tree information induced by MTT in previous works~\citep{liu2018learning,guo2020learning,nan2020reasoning}, unfortunately, the MTT would still produce arbitrary trees which inappropriate for the specific task if there is no structure supervision.
% gives us some random tree structures if we do not provide proper guidance during training. 
% the single application of MTT leads to a random tree structure. % Maybe an algorithm/ *figure can be added to illustrate
% To address these issues, we construct a multi-layer tree inducer which iteratively infers latent tree $N$ times. In contrast to multi-layer neural network like GCN~\citep{kipf2016semi} or Transformer~\citep{vaswani2017attention} where contextual representations are updated at every layer based on the output of previous layers, we only refine the tree structure during each iteration.
% To address these issues, we introduce an explicit token-level supervision to indicating each word in aspect term to constrain the root of the latent tree learned by the MTT must be the aspect word. 
% To address this issue, we introduce an explicit token-level supervision to indicate that each word in the aspect term should be assigned as the root. 
% To address this issue, we introduce an explicit token-level supervision indicating that the aspect words are identified as each aspect term which is used to constrain the root of the latent tree and which is also learned by MTT.
% Under the assumption that dependency relations with direct connections to an aspect may assist a model to focus more on related opinion words, 
\textcolor{black}{Under the assumption that inducing tree structures whose roots are within aspect term enables the model to better correlate the aspect and opinion words than the standard parse tree, we proposed to impose a soft constraint to encourage the aspect words $w \in \bm{a}$ to serve as the root of tree structure induced by MTT.
% we proposed to impose constraint to the root probability of aspect word $w_i \in \bm{a}$, making it more likely to be the root of tree structure induced by MTT.
}
% To address this issue, we introduce an explicit word-level supervision indicating that the aspect words are identified as the root of the latent tree.

\textcolor{black}{
Specifically, we introduce a cross-entropy loss for this assumption:
\begin{equation}
 \begin{aligned}
 \mathcal L_{a} =& -\sum_{i=1}^L\big (t_{i}\log(\boldsymbol{P}_{i}^{r}) \\ 
 	&+(1-t_i)\log(1-\boldsymbol{P}_{i}^{r}) \big )\text{,} 
 \end{aligned}
 \label{equ:rr}
\end{equation}
% \begin{equation}
%  \mathcal L_{a} = -\sum_{i=1}^L\big (t_{i}\log(\boldsymbol{P}_{i}^{r})+(1-t_i)\log(1-\boldsymbol{P}_{i}^{r}) \big )\text{,} 
% \end{equation}
where $t_i \in \{0, 1\}$ indicates whether the $i$-th token is the aspect word, $\bm{P}_i^{r}$ is the probability of the $i$-th token being the root from Equation \ref{equ:rootprob}. 
The nice property of this loss is that minimizing the loss is essentially adjusting the aspect words to be the root in the latent trees.
On the other hand, this supervision reduce the search space of inferring root for MTT in the training process.
}

% \textcolor{blue}{ \textbf{[TO BE REMOVED]}
% % Specifically, we introduce a binary label $y \in \{0, 1\}$ to indicate if a word is the aspect word. 
% On the other hand, we assign the probability of $y=1$ as the probability of the word being the root $\bm{P}_i^{r}$ (Equation \ref{equ:rootprob}).
% It is equivalent to say that we adjust the aspect words to be the root in the latent trees during the training process.
% As it is common for the aspect term to have multiple words in our task, we compute the probability of each aspect word to become the root.
% Thus, the roots of the trees induced by MTT could only be the aspect words rather than the others.
% In other words, we reduce the search space of inferencing root for MTT in the training process.
% Finally, the root refinement can be achieved by the following cross-entropy loss:
% % The idea is to predict a label for each token specifying whether it should be included in the aspect term. Given the root probabilities of the top layer of tree inducer $\boldsymbol{P}_{i}^{r} \in \mathbb R^{1 \times n}$, the aspect loss can be defined as:
% \begin{equation}
%  \begin{aligned}
%  \mathcal L_{a} =& -\sum_{n=1}^N\sum_{i=1}^L\big (y_{i}^{(n)}\log(\boldsymbol{P}_{i}^{r}) \\
%  	&+(1-y_i^{(n)})\log(1-\boldsymbol{P}_{i}^{r}) \big )\text{,} 
%  \end{aligned}
% \end{equation}
% %%\big, \Big \BIG
% where $y_i^{(n)} \in \{0, 1\}$ indicates whether the $i$-th token in $n$-th sentence is the aspect word, $L$ is the length of the sentence, and $N$ is the number of sentences. 
% }

Intuitively, the tree inducer module produces a random structure at early iterations during training since information propagates mostly between neighboring nodes.
As the roots are adjusted to the aspect words and the structure gets more refined when the loss becomes smaller, the tree inducer is more likely to generate an aspect-centric latent structure.
Our experiment in Section~\ref{root_refine} shows that the root refinement loss (Equation \ref{equ:rr}) is able to successfully guide the inducing of latent trees, in which the aspect word is consistent with its root.
% need to improve
% \paragraph{Tree Encoder}
\subsection{Tree Encoder}
Given contextualized representation $\bm{h}$ and the corresponding aspect-centric graph $\bm{P}$, we follow~\citet{kim2017structured} and~\citet{liu2018learning} to encode the tree information by structure attention mechanism:
% the tree information can be encoded by:
\begin{equation}
% \vspace{-1mm}
\begin{aligned}
\boldsymbol{s}_{i}^{p} &=\sum_{k=1}^{n} \boldsymbol{P}_{k i} \boldsymbol{h}_{k}+\boldsymbol{P}_{i}^{r} \boldsymbol{h}_{a} \\
\boldsymbol{s}_{i}^{c} &=\sum_{k=1}^{n} \boldsymbol{P}_{i k} \boldsymbol{h}_{i} \\
\boldsymbol{s}_{i} &=\tanh \left(\boldsymbol{W}_{s}\left[\boldsymbol{s}_{i}^{p}, \boldsymbol{s}_{i}^{c}, \boldsymbol{h}_{i}\right]\right)\text{,}
\end{aligned}
% \vspace{-1mm}%%cr version
\end{equation}
where $\boldsymbol{s}_{i}^{p} \in \mathbb R^d$ is the context representation gathered from possible parents of $\bm{h}_i$, $\boldsymbol{s}_{i}^{c}  \in \mathbb R^d$ is the context representation gathered from possible children, and $\boldsymbol{h}_{a}$ is the representation for the root node. We concatenate $\boldsymbol{s}_{i}^{p}$, $\boldsymbol{s}_{i}^{c}$ with $\boldsymbol{h}_{i}$ and transform with weights $\boldsymbol{W}_{s} \in \mathbb R^{d \times 3d}$ to obtain the structured representation of the $i$-th word $\boldsymbol{s}_i$.

% The final representation for classification is obtained by a pooling operation over the aspect words:
% \begin{equation}
% \bm{s}= pooling(\{\bm{s}_1,\bm{s}_2,...\bm{s}_A\})\text{,}
% \end{equation}
% where $pooling(\cdot)$ is a pooling function applied over the structured representation, $A$ is the number of words inside the aspect term, $\bm{s}$ is the aspect-specific structured representation.
% \paragraph{Classifier}
\subsection{Classifier}
Following~\citet{xu2019bert} and~\citet{sun2019utilizing}, we leverage $\bm{s}_0$, which is the structured aspect-aware representation of each sentence, to compute the probability over the different sentiment polarities as: 
% \vspace{-2mm}%%cr version
\begin{equation}
% \vspace{-1mm}%%cr version
y_p=\operatorname{softmax}\left(\bm{W}_{p} \bm{s}_0+\bm{b}_{p}\right) \text{,}
\end{equation}
where $\bm{W}_{p}$ and $\bm{b}_{p}$ are model parameters for the classifier, and $y_p$ is the predicted sentiment probability distribution.

The objective of the classifier is to minimize the cross-entropy loss for an instance ($\mathbf{x}, y$):
\begin{equation}
% \vspace{-2mm}%%cr version
\mathcal L_{s}=-\log P (y | \mathbf{x})
\end{equation}
% \begin{equation}
% % \vspace{-2mm}%%cr version
% \mathcal L_{s}=-\sum_{n=1}^{N} \sum_{c=1}^C y^{(n)} \log y_p^{(n)} \text{,}
% \end{equation}
% \begin{equation}
% % \vspace{-2mm}%%cr version
% \mathcal L_{s}=-\sum_{(\mathbf{x}, y) \in \mathcal{D}}\log P (y | \mathbf{x})
% \end{equation}
where $y \in \{positive, negative, neutral\}$.
% is the gold sentiment tag.
Our final objective function is a multi-task learning objective, defined as weighted sum of the loss on root refinement and classification:
\begin{equation}
\mathcal L=\alpha \mathcal L_{a} +(1-\alpha) \mathcal L_{s}\text{,}
\end{equation}
where $\alpha \in (0,1)$ is a coefficient that balances the contribution of each component in the training process. The hyper-parameter $\alpha$ is selected based on the performance on the validation set.

%%%exp
\section{Experiments} % todo 
\subsection{Experimental Setup}
We evaluate our proposed ACLT model on five benchmark datasets: the Laptop (Lap14) and Restaurant (Rest14) review datasets from SemEval 2014 Task4~\citep{Pontiki2014SemEval2014T4}, the Restaurant15 (Rest15) review dataset from SemEval 2015 Task12~\citep{Pontiki2015SemEval2015T1}, the Restaurant16 (Rest16) review dataset from SemEval 2016 Task5~\citep{Pontiki2016SemEval2016T5}, and Twitter posts from~\citep{dong2014adaptive}. Following the previous works~\citep{tang2016aspect,chen2017recurrent,wang2018learning}, we remove a few examples that have conflicting labels. 
\begin{table}[t!]
  \centering
  \resizebox{1\linewidth}{!}{
  \setlength{\tabcolsep}{0.8mm}{
    \begin{tabular}{lccccccccc}
    \toprule
    \multicolumn{1}{c}{\multirow{2}[4]{*}{\textbf{Dataset}}} & \multicolumn{3}{c}{\textbf{Train }} & \multicolumn{3}{c}{\textbf{Dev}} & \multicolumn{3}{c}{\textbf{Test}} \\
\cmidrule(lr){2-4} \cmidrule(lr){5-7} \cmidrule(lr){8-10}   & \#Pos. & \#Neu. & \#Neg. & \#Pos. & \#Neu. & \#Neg. & \#Pos. & \#Neu. & \#Neg. \\
    \midrule
    Lap14 & \textcolor{white}{0,}895   & \textcolor{white}{0,}418   & \textcolor{white}{0,}783   & \textcolor{white}{0}99    & \textcolor{white}{0}46    & \textcolor{white}{0}87    & 341   & 169   & 128 \\
    Rest14 & 1,948  & \textcolor{white}{0,}573   & \textcolor{white}{0,}726   & 216   & \textcolor{white}{0}64    & \textcolor{white}{0}81    & 728   & 196   & 196 \\
    Rest15 & \textcolor{white}{0,}821   & \textcolor{white}{0,0}32    & \textcolor{white}{0,}230   & \textcolor{white}{0}91    & \textcolor{white}{00}4     & \textcolor{white}{0}26    & 326   & \textcolor{white}{0}34    & 182 \\
    Rest16 & 1,116  & \textcolor{white}{0,0}62    & \textcolor{white}{0,}395   & 124   & \textcolor{white}{00}7     & \textcolor{white}{0}44    & 469   & \textcolor{white}{0}30    & 117 \\
    Twitter & 1,405  & 2,814  & 1,404  & 156   & 313   & 156   & 173   & 346   & 173 \\
    \bottomrule
    \end{tabular}}}%
    % \vspace{-3mm}%%cr version
    \caption{Statistics of datasets.}
    % \vspace{-3mm}%%cr version
  \label{tab:stat}
\end{table}%
We randomly split 10\% of data from the training dataset as the development dataset, and the model is only trained with the remaining data. 
Detailed statistics of the datasets can be found in Table~\ref{tab:stat}. 
All hyper-parameters are tuned based on the development set\footnote{We list some of the important hyper-parameters in Appendix~\ref{app:B}.  }. 
We employed the uncased version of the BERT-base~\citep{devlin2019bert} model in  PyTorch~\citep{wolf-etal-2020-transformers}\footnote{\url{https://github.com/huggingface/transformers}}. 
Following previous conventions, we repeat each experiment three times and average the results, reporting accuracy (Acc.) and macro-f1 ($F_1$).

\subsection{Baselines}

\textcolor{black}{The state-of-the-art baselines selected for comparison fall into three main categories: Syntax information free models,} dependency parse tree based models, and latent tree based models. Syntax information free models include: 
\squishlist
\item \textbf{TNet-AS}~\citet{li2018transformation}  implements a context-preserving mechanism to get the aspect-specific representations.
	\item \textbf{BERT-PT}~\citet{xu2019bert} explores a novel post-training approach on BERT to enhance the performance of BERT which has been fine-tuned for ABSA and RRC.
	\item \textbf{BERT-PAIR}~\citet{sun2019utilizing} constructs auxiliary sentences from the aspect and converts the ABSA task to a sentence-pair classification task.
	\item \textbf{BERT-SRC}~\cite{devlin2019bert} is the vanilla BERT model which directly uses the last layer's {\tt [CLS]} representation of the model as a classification feature.
\squishend

% ASGCN~\cite{zhang2019aspect}, CDT~\citep{sun2019aspect}, BiGCN~\citep{zhang2020convolution}, TNet-AS~\cite{li2018transformation}.

% As BERT outperforms existing open-source ABSA baselines by a large margin, we do not intend to exhaust existing implementations but instead focus on BERT-based baselines. 

The dependency parse tree based models are:
\squishlist
    \item \textbf{ASGCN}~\citet{zhang2019aspect} uses GCNs to capture the long-range dependencies between words.
    \item \textbf{CDT}~\citet{sun2019aspect} uses GCNs to integrate dependency parse tree information.
    \item \textbf{BiGCN}~\citet{zhang2020convolution} uses syntactic graph and lexical graph  to capture the global word co-occurrence information.
	\item \textbf{ASGCN+BERT} is a baseline that uses BERT instead of BiLSTM as the context encoder of ASGCN~\cite{zhang2019aspect}.
    % \item  \textbf{DepGCN+BERT} is a baseline that uses vanilla BERT as a context encoder and followed by a GCN to encode dependency parse tree. 
	\item  \textbf{R-GAT+BERT}~\cite{wang2020relational} is a dependency tree based model that greedily reshapes the dependency parse tree using manually defined rules.
\squishend

A latent tree based model:
\squishlist
	\item \textbf{KumaGCN+BERT}~\citet{chen2020inducing} constructs syntactic information by developing a gate mechanism to combine HardKuma structure and dependency parse tree.  
\squishend

We reproduce the results for baselines whenever the authors provide the source code. 
For ASGCN+BERT and KumaGCN+BERT models where the code is not made available as of this writing, we implement them by ourselves using the optimal hyper-parameter setting reported in their paper.
Since we randomly split 10\% of data from the training dataset as the development dataset, and the model is only trained with the remaining data, the results of R-GAT+BERT~\citep{wang2020relational} and KumaGCN+BERT~\citep{chen2020inducing}  are lower than which reported in the original paper.
In our experiments,  we report the average result and the mean absolute deviations over three runs with the random initialization. We stop training when iterations reached the maximum of 30 epochs.
\begin{table*}[htbp]
  \centering
  \resizebox{1\linewidth}{!}{
    \begin{tabular}{lccccccccccc}
    \toprule
    \multicolumn{1}{c}{\multirow{2}[4]{*}{\textbf{Models}}} & \multirow{2}[4]{*}{\textbf{Tree}} & \multicolumn{2}{c}{\textbf{Lap14}} & \multicolumn{2}{c}{\textbf{Rest14}} & \multicolumn{2}{c}{\textbf{Rest15}} & \multicolumn{2}{c}{\textbf{Rest16}} & \multicolumn{2}{c}{\textbf{Twitter}} \\
\cmidrule{3-12}          &       & Acc   & $F_1$   & Acc   & $F_1$    & Acc   & $F_1$   & Acc   & $F_1$    & Acc   & $F_1$ \\
    \midrule
    {TNet-AS$^\natural$} & {None} & 76.54 & {71.75} & 80.69 & {71.27} & -     & {-} & -     & {-} & 74.97 & 73.60 \\
    {BERT-PT$^\natural$} & {None} & 78.07 & {75.08} & 84.95 & {76.96} & -     & {-} & -     & {-} & -     & - \\
    {BERT-PAIR$^\natural$} & {None} & 78.99 & {75.03} & 84.46 & {76.98} & -     & {-} & -     & {-} & -     & - \\
    {BERT-SRC} & {None} & 77.59$\pm$0.18\textcolor{white}{$^\dag$} & {72.27$\pm$0.02\textcolor{white}{$^\dag$}} & 85.27$\pm$0.28\textcolor{white}{$^\dag$} & {77.61$\pm$0.38\textcolor{white}{$^\dag$}} & 81.73$\pm$0.45\textcolor{white}{$^\dag$} & {\underline{66.22$\pm$0.43}\textcolor{white}{$^\dag$}} & 90.91$\pm$0.07\textcolor{white}{$^\dag$} & {\underline{76.29$\pm$0.76}\textcolor{white}{$^\dag$}} & 73.12$\pm$0.29\textcolor{white}{$^\dag$} & 72.29$\pm$0.25\textcolor{white}{$^\dag$} \\
    {ASGCN$^\natural$} & {Dependency} & 75.55 & {71.05} & 80.77 & {72.02} & 79.89 & {61.89} & 88.99 & {67.48} & 72.15 & 70.40 \\
    {CDT$^\natural$} & {Dependency} & 77.19 & {72.99} & 82.30 & {74.02} & -     & {-} & 85.58 & {69.93} & 74.66 & 73.66 \\
    {BiGCN$^\natural$} & {Dependency} & 74.59 & {71.84} & 81.97 & {73.48} & 81.16 & {64.79} & 88.96 & {70.84} & 74.16 & 73.35 \\
    % {ASGCN+BERT$^*$} & {Dependency} & 78.00$\pm$0.24\textcolor{white}{$^\dag$} & {73.56$\pm$0.22\textcolor{white}{$^\dag$}} & 85.33$\pm$0.20\textcolor{white}{$^\dag$} & {77.59$\pm$0.06\textcolor{white}{$^\dag$}} & 83.03$\pm$0.25\textcolor{white}{$^\dag$} & {68.42$\pm$0.07\textcolor{white}{$^\dag$}} & 90.96$\pm$0.25\textcolor{white}{$^\dag$} & {74.92$\pm$0.31\textcolor{white}{$^\dag$}} & 73.99$\pm$0.19\textcolor{white}{$^\dag$} & 73.26$\pm$0.26\textcolor{white}{$^\dag$} \\
    {ASGCN+BERT} & {Dependency} & 77.90$\pm$0.10\textcolor{white}{$^\dag$} & {73.01$\pm$0.14\textcolor{white}{$^\dag$}} & 83.78$\pm$0.22\textcolor{white}{$^\dag$} & {75.02$\pm$0.51\textcolor{white}{$^\dag$}} & 80.69$\pm$045\textcolor{white}{$^\dag$} & {62.02$\pm$0.39\textcolor{white}{$^\dag$}} & 89.99$\pm$0.58\textcolor{white}{$^\dag$} & {74.46$\pm$0.16\textcolor{white}{$^\dag$}} & 72.78$\pm$0.71\textcolor{white}{$^\dag$} & 71.76$\pm$0.64\textcolor{white}{$^\dag$} \\
    {R-GAT+BERT$^*$} & {Dependency} & 78.53$\pm$0.31\textcolor{white}{$^\dag$} & {74.63$\pm$0.35\textcolor{white}{$^\dag$}} & \underline{85.63$\pm$0.24}\textcolor{white}{$^\dag$} & {\textbf{78.82$\pm$0.54}\textcolor{white}{$^\dag$}} & 81.61$\pm$0.78\textcolor{white}{$^\dag$} & {65.30$\pm$0.22\textcolor{white}{$^\dag$}} & \underline{90.96$\pm$0.18}\textcolor{white}{$^\dag$} & {75.26$\pm$0.39\textcolor{white}{$^\dag$}} & 73.80$\pm$0.61\textcolor{white}{$^\dag$} & 72.63$\pm$0.46\textcolor{white}{$^\dag$} \\
    {KumaGCN+BERT$^\ddag$} & {Latent} & \underline{79.57$\pm$0.28}\textcolor{white}{$^\dag$} & {\underline{75.61$\pm$0.28}\textcolor{white}{$^\dag$}} &  84.91$\pm$0.30\textcolor{white}{$^\dag$}     & {77.22$\pm$0.37\textcolor{white}{$^\dag$}} & \underline{82.10$\pm$0.62}\textcolor{white}{$^\dag$} & {65.56$\pm$0.61\textcolor{white}{$^\dag$}} & 90.80$\pm$0.47\textcolor{white}{$^\dag$} & {74.93$\pm$0.97\textcolor{white}{$^\dag$}} & \underline{74.33$\pm$0.32}\textcolor{white}{$^\dag$}      & 
    \underline{73.42$\pm$0.31}\textcolor{white}{$^\dag$}\\
    \midrule
   {ACLT} & {Latent} & \textbf{79.68$\pm$0.38\textcolor{white}{$^\dag$}} & \textbf{75.83$\pm$0.03\textcolor{white}{$^\dag$}} & \textbf{85.71$\pm$0.06\textcolor{white}{$^\dag$}} & \underline{78.44$\pm$0.09}\textcolor{white}{$^\dag$} & \textbf{84.44$\pm$0.08}$^\dag$ & \textbf{72.08$\pm$0.08}$^\dag$ &\textbf{ 92.15$\pm$0.14}$^\dag$ & \textbf{78.64$\pm$0.19}$^\dag$ &\textbf{ 75.48$\pm$0.16}$^\dag$ & \textbf{74.51$\pm$0.32}$^\dag$ \\
    \bottomrule
    \end{tabular}}%
    %   \vspace{-3mm}%%cr version
    %   \caption{Main Results (\%). The results with the symbol~$\natural$ are computed based on their open implementations. Models with the * symbol are equipped with the external parse tree. The~$\dag$ marker refers to $p$-value $<$ 0.05 of the one-tailed paired t-test compared to the second-best results.} 
            \caption{Main Results (\%). The results of model with the symbol~$\natural$ are retrieved from the original paper, and those with the * symbol are computed based on their open implementations. $\ddag$ denotes the model using both the dependency parse tree and the latent tree. The best results on each dataset are in bold. The second-best ones are underlined. The~$\dag$ marker refers to $p$-value $<$ 0.05 in comparison with the second-best results.} 
    %  \vspace{-2mm}%%cr version
  \label{tab:mainresult}%
\end{table*}%
\subsection{Main Results}
\textcolor{black}{As shown in Table \ref{tab:mainresult}, dependency tree based models and latent tree based models generally achieve better results than syntax information free models, suggesting that syntactic information indeed benefits the ABSA task and enables it to achieve promising results.}

\textcolor{black}{Our model consistently outperforms the models which do not use any syntactic information.}
For example, ACLT improves upon the BERT-SRC model by 3.56 points in terms of $F_1$ on the Lap14 dataset, which suggests that our proposed model is able to induce an effective latent tree for ABSA in an end-to-end fashion.
In particular, with the exception of R-GAT+BERT on the Rest14 dataset in terms of $F_1$, our model surpassed all compared models by a significant margin. 
For example, our model achieves 72.08 and 78.64 $F_1$ on the Rest15 and Rest16 datasets, which significantly outperform the current state-of-the-art model KumaGCN+BERT, under the same setting. 
The statistics empirically show that compared to the models that use syntactic information, ACLT can induce a more informative latent task-specific structure to establish effective connections between aspect words and context.
Our ACLT model also shows its superiority over all baselines in terms of accuracy.

% Using a BERT-Large version pre-train embedding, we observed that our model ACLT-BERT-L outperforms all the baselines on all the metrics except kumaGCN+BERT on the Lap14 and Twitter datasets. In addition, ACL-BERT-L significantly outperforms the strong baseline BERT-L-SRC with $p<$0.01 on all the datasets. This result indicates that aspect-centric trees can be flexibly adapted to the BERT-Large in order to capture richer non-local dependency structures for the ABSA task. 
%  Furthermore, we may deduce that our proposed method can also outperform  other powerful pre-trained language models.
\begin{table}[t!]
  \centering
 \resizebox{1\linewidth}{!}{
 \setlength{\tabcolsep}{3mm}{
    \begin{tabular}{lccccc}
    \toprule
    & \multicolumn{5}{c}{\textbf{Positive opinion words}} \\
    Tree  & \textit{great} & \textit{good}  & \textit{excellent} & \textit{fresh} & \textit{delicious} \\
    \midrule
    \textbf{Parser} & 4.38     & 4.50   & 5.11   & 8.02   & 6.91 \\
    \textbf{MTT}  & 3.84   & 4.47   & 5.05   & 6.61   & 4.43 \\
    \textbf{ACLT}  & \textbf{2.81}   & \textbf{3.40}   & \textbf{4.08}    & \textbf{4.84}   & \textbf{3.57} \\
    \midrule\midrule
    & \multicolumn{5}{c}{\bf Negative opinion words} \\
    Tree  & \textit{rude} & \textit{small}  & \textit{bad} & \textit{awful} & \textit{worst} \\
    \midrule
    \textbf{Parser} & 6.67   & 11.27   & 9.44   & 4.00   & 3.88 \\
    \textbf{MTT}  & 5.87   & 10.18   & 8.56    & 4.00   & 3.75 \\
    \textbf{ACLT}  & \textbf{3.27}   &\textcolor{white}{0}\textbf{6.45}   & \textbf{3.89}    & \textbf{2.89}   & \textbf{3.13} \\
    \bottomrule
    \end{tabular}}}%
    % \vspace{-3mm}%%cr version
     \caption{The average distances (lower is better) between the top five opinion words and aspect words. }
    %  \vspace{-3mm}%%cr version
  \label{tab:polarity_stat}%
\end{table}%
\subsubsection*{Does ACLT shorten the distances between aspect and opinion words?}
To gain further insight on the relationship between aspect and opinion words in the text, we inspect the distance just between aspect words and selected opinion words. Specifically, we first selected the top five most frequent positive and negative opinion words, respectively, in the Rest14 dataset. We define the distance between the aspect and opinion words to be the number of interaction hops between them. Thus we can calculate the distance between these opinion words and aspect words in a parse tree\footnote{We use Chu-Liu-Edmonds' algorithm to extract the aspect-centric trees. More detail can be found in section~\ref{sec:cs}.} and an aspect-centric tree, respectively. %\footnote{We use Chu-Liu-Edmonds' algorithm~\citep{edmonds1967optimum} to extract the aspect-centric trees, more detail can be found in section~\ref{sec:cs}.}

Table~\ref{tab:polarity_stat} presents various statistics for the average distance of aspect and opinion words in the trees produced by spaCy dependency parser (\textbf{Parser}), the Matrix Tree Theory without specific root refinements (\textbf{MTT}), and our model (\textbf{ACLT}). 
As can be seen,  in our aspect-centric latent tree, the average distance between opinion words and aspect words is shorter than those in dependency parse tree and MTT. 
We also observe that without the root refinement, the average distance between opinion words and aspect words in MTT is roughly equivalent to the parse tree.
% This demonstrates that our ACLT model is able to establish relationships  between aspect and opinion words by promoting the interaction hops between.
% These results confirm our hypothesis that inducing tree structures whose roots are within aspect term in the ABSA task enables the model to correlate the aspect and opinion words better.
\textcolor{black}{These results confirm our hypothesis that inducing tree structures whose roots are within aspect term enables the model to better correlate the aspect and opinion words than the standard parse tree.}
\begin{table*}[t!]
  \centering
    \resizebox{0.8\linewidth}{!}{
    \setlength{\tabcolsep}{3mm}{
    \begin{tabular}{lcccccccccc}
    \toprule
    \multicolumn{1}{c}{\multirow{2}[4]{*}{\textbf{Models}}} & \multicolumn{2}{c}{\textbf{Lap14}} & \multicolumn{2}{c}{\textbf{Rest14}} & \multicolumn{2}{c}{\textbf{Rest15}} & \multicolumn{2}{c}{\textbf{Rest16}} & \multicolumn{2}{c}{\textbf{Twitter}} \\
\cmidrule(lr){2-3} \cmidrule(lr){4-5} \cmidrule(lr){6-7} \cmidrule(lr){8-9} \cmidrule(lr){10-11}        & Acc.   & $F_1$     & Acc.   &  $F_1$     & Acc.   &  $F_1$     & Acc.   &  $F_1$     & Acc.   &  $F_1$  \\
    \midrule
    BERT-SRC  & 77.6 & 72.3 & 85.3 & 77.6 & 81.7 & 66.2 & 90.9 & 76.3 & 73.1 & 72.3 \\
    Parser+GCN & 78.0 & 73.6 & 85.3& 77.6& 83.0 & 68.4 & 91.0 & 74.9 & 74.0 & 73.3 \\
    MTT+GCN & 78.9 & 74.7 & 84.7 & 76.3 & 81.9 & 67.6 & 91.2 & 74.8 & 75.3 & 74.3 \\
    Kuma+GCN & 78.1 & 73.5 & 85.3 & 77.9 & 80.3 & 63.2 & 90.4 & 75.2 & 74.6 & 73.7 \\
     \midrule
    ACLT+GCN & 78.6 & 74.3 & \textbf{86.3} & \textbf{79.4} & 83.1 & 70.2 & 91.8 & 76.7 & \textbf{75.6} & \textbf{74.7} \\
    ACLT  & \textbf{79.7} & \textbf{75.8} & 85.7 & 78.4 & \textbf{84.4} & \textbf{72.1} & \textbf{92.2} & \textbf{78.6} & 75.5 & 74.5 \\
    \bottomrule
    \end{tabular}}}
    % }}%
    % \vspace{-3mm}%%cr version
      \caption{The performance of BERT with our aspect-centric latent tree vs. BERT with other tree structures.}
  \label{tab:aclt_parser}%
%   \vspace{-1mm}%%cr version
\end{table*}
%单列
% \begin{table}[t!]
%   \centering
%     \resizebox{1\linewidth}{!}{
%     \setlength{\tabcolsep}{1.5mm}{
%     \begin{tabular}{lcccccccccc}
%     \toprule
%     \multicolumn{1}{c}{\multirow{2}[4]{*}{\textbf{Models}}} & \multicolumn{2}{c}{\textbf{Lap14}} & \multicolumn{2}{c}{\textbf{Rest14}} & \multicolumn{2}{c}{\textbf{Rest15}} & \multicolumn{2}{c}{\textbf{Rest16}} & \multicolumn{2}{c}{\textbf{Twitter}} \\
% \cmidrule(lr){2-3} \cmidrule(lr){4-5} \cmidrule(lr){6-7} \cmidrule(lr){8-9} \cmidrule(lr){10-11}        & Acc.   & $F_1$     & Acc.   &  $F_1$     & Acc.   &  $F_1$     & Acc.   &  $F_1$     & Acc.   &  $F_1$  \\
%     \midrule
%     BERT-SRC  & 77.6 & 72.3 & 75.3 & 77.6 & 81.7 & 66.2 & 90.9 & 76.3 & 73.1 & 72.3 \\
%     Parser+GCN & 78.0 & 73.6 & 85.3& 77.6& 83.0 & 68.4 & 91.0 & 74.9 & 74.0 & 73.3 \\
%     MTT+GCN & 78.9 & 74.7 & 84.7 & 76.3 & 81.9 & 67.6 & 91.2 & 74.8 & 75.3 & 74.3 \\
%     Kuma+GCN & 78.1 & 73.5 & 85.3 & 77.9 & 80.3 & 63.2 & 90.4 & 75.2 & 74.6 & 73.7 \\
%      \midrule
%     ACLT+GCN & 78.6 & 74.3 & \textbf{86.3} & \textbf{79.4} & 83.1 & 70.2 & 91.8 & 76.7 & \textbf{75.6} & \textbf{74.7} \\
%     ACLT  & \textbf{79.7} & \textbf{75.8} & 85.7 & 78.4 & \textbf{84.4} & \textbf{72.1} & \textbf{92.2} & \textbf{78.6} & 75.5 & 74.5 \\
%     \bottomrule
%     \end{tabular}
%     }}%
%     % \vspace{-3mm}%%cr version
%       \caption{The performance of BERT with our aspect-centric latent tree vs. BERT with explicit parse tree.}
%   \label{tab:aclt_parser}%
% %   \vspace{-1mm}%%cr version
% \end{table}
\subsection{Model Analysis}
% In this section,  we conduct qualitative and quantitative analysis to demonstrate the effectiveness of our aspect-centric latent tree.
\subsubsection*{Effect of different tree representations}
% \paragraph{Aspect-Centric Latent Tree or Parse Tree?}
Our proposed aspect-centric latent tree, the latent Matrix tree, and the standard dependency parse tree all represent the structure of a sentence. 
Nevertheless, the differences between them and how they directly affect the aspect-based results need to be further investigated. 
In this section, we first use BERT-base as a contextual encoder, then use GCN to encode dependency parse tree information (Parser+GCN), latent Matrix tree information (MTT+GCN), latent Kuma structure (Kuma+GCN\footnote{For a fair comparison, we only use the Kuma structure rather than combining the dependency tree and the Kuma structure in this experiment.}) and our aspect-centric tree information (ACLT+GCN).
Table~\ref{tab:aclt_parser} summarizes the results.

We observe that models incorporated with syntactic information generally outperform the vanilla BERT-SRC, indicating that syntactic information benefits the ABSA task. 
\textcolor{black}{Such a phenomenon can also be observed in other fundamental NLP tasks~\cite{jie2019dependency,xu2021better}.}
Moreover, we also found that both our ACLT and ACLT+GCN model consistently outperform models equipped with other dependency trees by a significant margin.
These results demonstrate that the aspect-centric tree induced by our model is indeed capable of effectively building relationships between aspect and context words for the ABSA task.
Under the same setting, ACLT+GCN outperforms Parser+GCN, MTT+GCN, and Kuma+GCN on all the datasets.
\textcolor{black}{In particular, our ACLT+GCN obtains 1.8, 2.6, and 7 points improvement over Parser+GCN, MTT+GCN, and Kuma+GCN on Rest15 in terms of $F_1$.}
Moreover, ACLT+GCN outperforms ACLT on the Rest14 and the Twitter datasets, indicating using a GCN as a tree encoder can boost the model performance to a certain extent. 

We also have similar observations for our ACLT model under the setting of accuracy.
These experimental results demonstrate that our proposed aspect-centric latent tree is a more effective structure for ABSA, compared to the parse tree. 
Interestingly, we observe that BERT cannot achieve a promising result on all datasets when introduced with the parse tree structure. 
For example, Parser+GCN drops 1.4 points in $F_1$ on the Rest16 dataset in comparison with vanilla BERT-SRC.  
This suggests that a dependency parse tree structure may not be able to capture the complicated interactions between aspect and opinion words effectively.
% two rows table'

% \begin{table}[htbp]
%   \centering
%  \resizebox{1\linewidth}{!}{
%  \setlength{\tabcolsep}{3.5mm}{
%     \begin{tabular}{lccccc}
%     \toprule
%     Tree  & \textbf{rude} & \textbf{small}  & \textbf{bad} & \textbf{awful} & \textbf{worst} \\
%     \midrule
%     Parser & 51   & 75   & 31   & 14   & 12 \\
%     ACLT  & 23   & 48   & 22    & 9   & 7 \\
%     \bottomrule
%     \end{tabular}}}%
%      \caption{The sum of the distances between top five negative opinion words and aspect words}
%   \label{tab:neg_stat}%
% \end{table}%

% Table generated by Excel2LaTeX from sheet 'Analysis'
% \begin{table*}[htbp]
%   \centering
%   \resizebox{1\linewidth}{!}{
%   \setlength{\tabcolsep}{3.8mm}{
%     \begin{tabular}{lccccc|ccccc}
%     \toprule
%     \multicolumn{1}{c}{\multirow{2}[4]{*}{Tree}} & \multicolumn{5}{c|}{Positive opinion words} & \multicolumn{5}{c}{Negative opinion words} \\
% \cmidrule(lr){2-6}   \cmidrule(lr){7-11}       & great & good  & excellent & fresh & delicious & rude  & small & bad   & awful & worst \\
%     \midrule
%     Parser & 520   & 283   & 144   & 322   & 240   & 51    & 75    & 31    & 14    & 12 \\
%     ACLT  & 285   & 156   & 80    & 194   & 128   & 23    & 48    & 22    & 9     & 7 \\
%     \bottomrule
%     \end{tabular}%
%   \label{tab:polarity_stat}}}%
%   \caption{Add caption}
% \end{table*}%

% Table generated by Excel2LaTeX from sheet 'Analysis'

\subsubsection*{Did root refinement work?} 
\begin{figure}[t!]
% \vspace{-5mm}%%cr version
  \centering
  {\includegraphics[width=0.5\textwidth]{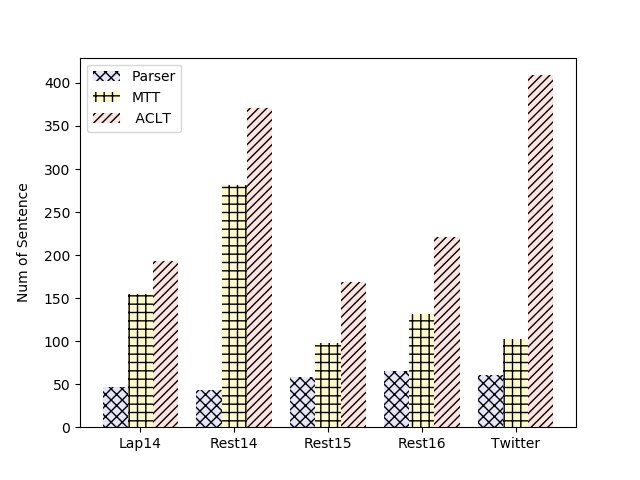}\label{vis(root_static)}}
%   \vspace{-4mm}%%cr version
  \caption{The number of sentences where the aspect words are roots under three different types of trees.}
%   \vspace{-4mm}%%cr version
  \label{fig:root_num}
\end{figure}
\label{root_refine}
% We investigate the extent to which the root refinement help to transfer the root to aspect words
We quantify the effectiveness of root refinement that adjusts the aspect words to be the root.
% We investigate the extent to which the root refinement help to transfer the root to aspect words
% enables to refine the root of induced tree. 
We experiment with three different structures, including the dependency parse tree obtained by spaCy (Parser), the tree directly induced by MTT without specific root refinements (MTT), and the aspect-centric tree induced by our model (ACLT). 
Figure~\ref{fig:root_num} shows the number of sentences where the aspect word is consistent with its root under three different tree structures in each dataset. 
Compared to the other two tree structures, we observe that the roots of our learned trees are consistent with the aspect words in most sentences. 
For example, in the Rest16 test dataset, there are 421 sentences in which the aspect words are consistent with the root words using the ACLT model. 
These results demonstrate that the problem of inconsistency between root and aspect has come close to being solved with our ACLT model.
% \subsubsection*{Does the entire tree matter?}
\begin{table}[t!]
  \centering
  \resizebox{1\linewidth}{!}{
    \begin{tabular}{lcccc}
    \toprule
    \multicolumn{1}{c}{\multirow{2}[4]{*}{\textbf{Models}}} & \multicolumn{2}{c}{Rest14} & \multicolumn{2}{c}{Twitter} \\
\cmidrule(lr){2-3} \cmidrule(lr){4-5}        & Acc.   &  $F_1$ & Acc.    & $F_1$ \\
    \midrule
    BERT-SRC & 85.27 & 77.61 & 73.12 & 72.29 \\
% \cmidrule{1-1}
\midrule ACLT (Entire Tree) & 85.71 & 78.44 & 75.48 & 74.51 \\
    Pruned Tree ($k=1$) & 85.27 & 77.37 & 74.56 & 73.75 \\
    Pruned Tree ($k=2$) &  84.91     &    77.05   & 73.84 & 72.72 \\
    \midrule R-GAT+BERT (Entire Tree) & 85.63 & 78.82 & 73.80 & 72.63 \\
    Pruned Tree ($k=1$) & 85.71 & 79.14 & 74.71& 73.85 \\
    Pruned Tree ($k=2$) &  84.73     &    78.67   & 73.99 & 73.16 \\
    \midrule KumaGCN+BERT (Entire Tree) & 84.91 & 77.22 & 74.33 & 73.42 \\
    Pruned Tree ($k=1$) & 84.91 & 76.73 & 75.14 & 73.90 \\
    Pruned Tree ($k=2$) &  85.09     &    77.35   & 75.43 & 74.42 \\
    \bottomrule
    \end{tabular}}%
    % \vspace{-3mm}%%cr version
     \caption{The results of ACLT, R-GAT+BERT and KumaGCN+BERT with different tree pruning. 
     $k$=1: only keep the first-order edges to the aspect. 
     $k$=2: keep both the first-order and second-order edges.
    %  k=1 and k=2 indicate the aspect-centric latent tree we used has been pruned the relations beyond the first-order and the second-order of aspect, respectively.
     }
  \label{tab:pruned tree}%
\end{table}%
\subsubsection*{Effect of tree pruning}
% To further investigate the entire aspect-centric tree, we examine our ACLT under different tree pruning.
\textcolor{black}{To further investigate the effect of different tree structures on model performance, we examine ACLT, R-GAT+BERT, and KumaGCN+BERT with different tree pruning.
More specifically, for R-GAT+BERT using the standard prase tree, we 
discard the dependency relation beyond first-order ($k$=1) and second-order ($k$=2) to aspects, respectively. 
Following ~\citet{guo2020learning}, we mask the information of the adjacency matrix $\boldsymbol{P}$ (Equation \ref{equ:adj}) that is beyond first-order ($k$=1) and second-order ($k$=2)  with respect to the aspect for KumaGCN+BERT and our ACLT model.}
As shown in Table~\ref{tab:pruned tree},  on the Twitter dataset, our ACLT yields the best performance with the entire tree, outperforming the first-order pruned tree and second-order pruned tree by 0.76 and 1.79 points in terms of $F_1$, respectively. This indicates it is necessary to induce an entire aspect-centric latent tree rather than its pruned subtree in our model.
Interestingly, we observe that R-GAT+BERT and KumaGCN+BERT achieve the best results in cases of Pruned Tree ($k=1$) and Pruned Tree ($k=2$), respectively.
It is likely because that both R-GAT+BERT and KumaGCN+BERT rely on the parse tree. Nevertheless, only a small part of the standard parse tree is related to the ABSA task. Introducing the entire tree may prevent the model from effectively capturing the relationships between opinion words and aspect words.

\subsubsection*{Ablation Study} % 
We conducted experiments to examine the effectiveness of the major components of our ACLT model, and Table~\ref{tab:Ablation} shows the ablation results on the five datasets we used. 
We observe that both latent tree and root refinement component contribute to the main model.
% , as performance deteriorates with removal of any component. 
\textcolor{black}{Specifically, with removal of the root refinement module, performance of ACLT drops considerably, leading to a 5.2 and 4.8 decrease, in terms of $F_1$, on the Rest15 dataset and the Rest16 dataset, respectively.}
This result illustrates that refining root to aspect words plays a crucial role in learning a task-specific latent structure for ABSA. 
The performance drop on fixed root indicates that computing each aspect word’s probability to become the root is essential for achieving good performance.
\begin{table}[t!]
  \centering
\resizebox{1\linewidth}{!}{
\setlength{\tabcolsep}{0.8mm}{
    \begin{tabular}{lcccccccccc}
    \toprule
    \multicolumn{1}{c}{\multirow{2}[4]{*}{\textbf{Models}}} & \multicolumn{2}{c}{\textbf{Lap14}} & \multicolumn{2}{c}{\textbf{Rest14}} & \multicolumn{2}{c}{\textbf{Rest15}} & \multicolumn{2}{c}{\textbf{Rest16}} & \multicolumn{2}{c}{\textbf{Twitter}} \\
\cmidrule(lr){2-3} \cmidrule(lr){4-5} \cmidrule(lr){6-7} \cmidrule(lr){8-9} \cmidrule(lr){10-11}          & Acc.   & $F_1$    & Acc.   & $F_1$    & Acc.   & $F_1$    & Acc.   & $F_1$    & Acc.   & $F_1$ \\
    \midrule
    % BERT-B-SRC & 78.06 & 73.28 & 84.91 & 77.05 & 82.66 & 67.55 & 90.91 & 74.91 & 73.69 & 72.76 \\
    % \midrule
    Full Model & 79.7 & 75.8 & 85.7 & 78.4 & 84.4 & 72.1 & 92.2 & 78.6 & 75.5 & 74.5 \\
        \textcolor{white}{---}w/o\textcolor{white}{-} Root Refinement & 77.8 & 73.2 & 84.5 & 76.4 & 81.5 & 66.9 & 89.8 & 73.8 & 73.4 & 72.6 \\
    \textcolor{white}{---}w/o\textcolor{white}{-} Latent Tree & 77.8 & 73.4 & 84.7 & 76.7 & 83.9 & 69.1 & 91.5 & 77.6 & 74.0 & 73.3 \\
    \textcolor{white}{---}w\textcolor{white}{/o}\textcolor{white}{-} Fixed Root & 79.2& 75.0 & 84.6 & 76.3 & 83.2 &68.9  & 91.1 & 75.0 & 75.0 & 74.1  \\
    \bottomrule
    \end{tabular}}}%
    % \vspace{-2mm}%%cr version
      \caption{Ablation study of ACLT on various datasets. w/o and w indicate without and with, resepectively. Fixed Root means the tree's root is fixed on the first word of aspect term~\citep{wang2020relational}.}
    % \vspace{-4mm}
    %   \footnote{We refer the interested reader to ~\citet{wang2020relational} for more details}}
  \label{tab:Ablation}%
\end{table}

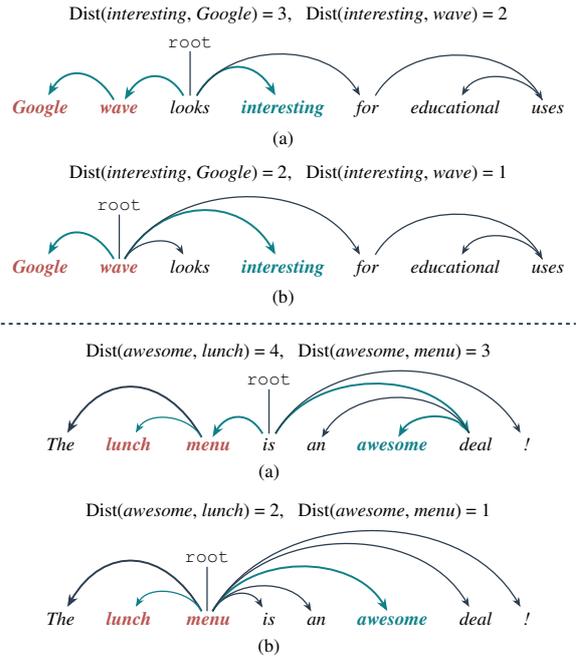
\begin{figure}[t!]
	\centering
	\adjustbox{max width=1\linewidth}{
		\begin{tikzpicture}[node distance=2.0mm and 2.0mm, >=Stealth, 
			wordnode/.style={draw=none, minimum height=5mm, inner sep=0pt},
			chainLine/.style={line width=1pt,-, color=fontgray},
			entbox/.style={draw=black, rounded corners, fill=red!20, dashed},
			invis/.style={draw=none, circle, inner sep=0pt, fill=white}
			]
			
			\matrix(sent1) [matrix of nodes, row sep=36mm, column sep=5mm, nodes in empty cells, execute at empty cell=\node{\strut};]
			{
				\textcolor{customred}{\textbf{\textit{\Large Google}}} & [1mm]\textcolor{customred}{\textbf{\textit{\Large wave}}} &[1mm]\textit{\Large looks}  & [1mm] \textcolor{customblue}{\textbf{\textit{\Large interesting}}}   &  [1mm] \textit{\Large  for }& [1mm] \textit{\Large educational} & [1mm] \textit{\Large uses}   \\
				\textcolor{customred}{\textbf{\textit{\Large Google}}} & [1mm]\textcolor{customred}{\textbf{\textit{\Large wave}}} &[1mm]\textit{\Large looks}  & [1mm]\textcolor{customblue}{\textbf{\textit{\Large interesting}}}    &  [1mm] \textit{\Large  for }& [1mm] \textit{\Large educational} & [1mm] \textit{\Large uses}   \\
			};

			\draw[chainLine, <-, color=customblue, line width=1.5pt] (sent1-1-1) to [out=60,in=120, looseness=1.4] node[above, yshift=-1mm, color=black]{} (sent1-1-2);
			\draw[chainLine, <-, customblue, line width=1.5pt] (sent1-1-2) to [out=60,in=120, looseness=1.4] node[above, yshift=-1mm, color=black]{} (sent1-1-3);
			
			\draw[chainLine, ->, customblue, line width=1.5pt] (sent1-1-3) to [out=60,in=120, looseness=1.4] node[above, yshift=-1mm, color=black]{} (sent1-1-4);
			\draw[chainLine, ->] (sent1-1-3) to [out=60,in=120, looseness=1] node[above, yshift=-1mm, color=black]{} (sent1-1-5);
			%			\draw[chainLine, ->] (sent1-1-4) to [out=60,in=120, looseness=1] node[above, yshift=-1mm, color=black]{} (sent1-1-8);
			\draw [chainLine, ->] (sent1-1-5) to [out=60,in=120, looseness=1] node[above, yshift=-1mm, xshift=-3mm, color=black]{} (sent1-1-7);
			\draw [chainLine, <-] (sent1-1-6) to [out=60,in=120, looseness=1] node[above, yshift=-1mm, xshift=-3mm, color=black]{} (sent1-1-7);

			\draw[chainLine, <-, color=customblue, line width=1.5pt] (sent1-2-1) to [out=60,in=120, looseness=1.4] node[above, yshift=-1mm, color=black]{} (sent1-2-2);
			\draw[chainLine, ->] (sent1-2-2) to [out=60,in=120, looseness=1] node[above, yshift=-1mm, color=black]{} (sent1-2-3);
			\draw[chainLine, ->, customblue, line width=1.5pt] (sent1-2-2) to [out=60,in=120, looseness=1.2] node[above, yshift=-1mm, color=black]{} (sent1-2-4);
			\draw[chainLine, ->] (sent1-2-2) to [out=60,in=120, looseness=1] node[above, yshift=-1mm, color=black]{} (sent1-2-5);
			\draw[chainLine, ->] (sent1-2-7) to [out=120,in=60, looseness=1] node[above, yshift=-1mm, color=black]{} (sent1-2-6);
			\draw[chainLine, ->] (sent1-2-5) to [out=60,in=120, looseness=1] node[above, yshift=-1mm, color=black]{} (sent1-2-7);
			%			\draw[chainLine, ->] (sent1-2-3) to [out=60,in=120, looseness=1] node[above, yshift=-1mm, color=black]{} (sent1-2-8);

			\node[wordnode, above= of sent1-1-3, yshift=10mm] (root1) {\Large \texttt{root}};
			\draw[chainLine, -] (root1) to  (sent1-1-3);
			
			\node[wordnode, above= of sent1-2-2, yshift=10mm] (root2) {\Large \texttt{root}};
			\draw[chainLine, -] (root2) to  (sent1-2-2);
			
% 			\node[wordnode, above= of sent1-1-1, xshift=4mm, yshift=14mm] (da) {\LARGE\texttt{Dist=5}};
% 			\node[wordnode, above= of sent1-2-1, xshift=4mm, yshift=16mm] (db) {\LARGE\texttt{Dist=3}};
			\node[wordnode, above= of sent1, yshift=16mm] (da) {\Large  {Dist(\textit{interesting}, \textit{Google}) = 3,~~  Dist(\textit{interesting}, \textit{wave}) = 2}};
			\node[wordnode, below= of da, yshift=-36mm] (db) {\Large  {Dist(\textit{interesting}, \textit{Google}) = 2,~~  Dist(\textit{interesting}, \textit{wave}) = 1}};
% 			{\large  {Dist(\textit{Loving}, \textit{harry}) = 4,~~  Dist(\textit{Loving}, \textit{potter}) = 3}}
			\node[wordnode, below=of sent1-1-4](a){\Large (a)};
			\node[wordnode, below=of sent1-2-4](b){\Large (b)};

			\node[invis, below =of sent1, yshift=-8mm](invismiddle){};
			\node[invis, left =of invismiddle, yshift=0mm, xshift=-75mm](invis1){};
			
			\node[invis, right =of invismiddle, yshift=0mm, xshift=75mm](invis2){};
			\draw[chainLine, -, dashed, line width=1.5pt, fontgray] (invis1) to (invis2);

			\matrix(sent2) [matrix of nodes, below=of invismiddle, yshift=-26mm, row sep=41mm, column sep=5mm, nodes in empty cells, execute at empty cell=\node{\strut};]
			{
				\textit{\Large The} & [1mm]\textcolor{customred}{\textbf{\textit{\Large lunch}}} &[1mm]\textcolor{customred}{\textbf{ \textit{\Large menu}}} & [1mm] \textit{\Large is}   &  [1mm] \textit{\Large  an }& [1mm] \textbf{\textcolor{customblue}{\textit{\Large awesome}}} & [1mm] \textit{\Large deal} & [1mm] \textit{\Large !} \\
				\textit{\Large The} & [1mm]\textcolor{customred}{\textbf{\textit{\Large lunch}}} &[1mm]\textcolor{customred}{\textbf{ \textit{\Large menu}}} & [1mm] \textit{\Large is}   &  [1mm] \textit{\Large  an }& [1mm]\textbf{\textcolor{customblue}{\textit{\Large awesome}}}  & [1mm] \textit{\Large deal} & [1mm] \textit{\Large !} \\
			};

			\draw[chainLine, ->, color=fontgray, line width=1.5pt] (sent2-1-3) to [out=120,in=60, looseness=1.4] node[above, yshift=-1mm, color=black]{} (sent2-1-1);
			\draw[chainLine, <-,customblue] (sent2-1-2) to [out=60,in=120, looseness=1] node[above, yshift=-1mm, color=black]{} (sent2-1-3);
			
			\draw[chainLine, <-, customblue, line width=1.5pt] (sent2-1-3) to [out=60,in=120, looseness=1.4] node[above, yshift=-1mm, color=black]{} (sent2-1-4);
			\draw[chainLine, ->, customblue, line width=1.5pt] (sent2-1-4) to [out=60,in=120, looseness=1] node[above, yshift=-1mm, color=black]{} (sent2-1-7);
			\draw[chainLine, ->] (sent2-1-4) to [out=60,in=120, looseness=1] node[above, yshift=-1mm, color=black]{} (sent2-1-8);
			\draw [chainLine, ->, customblue, line width=1.5pt] (sent2-1-7) to [out=120,in=60, looseness=1] node[above, yshift=-1mm, xshift=-3mm, color=black]{} (sent2-1-6);
			\draw [chainLine, ->] (sent2-1-7) to [out=120,in=60, looseness=1] node[above, yshift=-1mm, xshift=-3mm, color=black]{} (sent2-1-5);

			\draw[chainLine, ->, color=fontgray, line width=1.5pt] (sent2-2-3) to [out=120,in=60, looseness=1.4] node[above, yshift=-1mm, color=black]{} (sent2-2-1);
			\draw[chainLine, <-, customblue] (sent2-2-2) to [out=60,in=120, looseness=1] node[above, yshift=-1mm, color=black]{} (sent2-2-3);
			\draw[chainLine, ->] (sent2-2-3) to [out=60,in=120, looseness=1.2] node[above, yshift=-1mm, color=black]{} (sent2-2-4);
			\draw[chainLine, ->] (sent2-2-3) to [out=60,in=120, looseness=1] node[above, yshift=-1mm, color=black]{} (sent2-2-5);
			\draw[chainLine, ->, customblue, line width=1.5pt] (sent2-2-3) to [out=60,in=120, looseness=1] node[above, yshift=-1mm, color=black]{} (sent2-2-6);
			\draw[chainLine, ->] (sent2-2-3) to [out=60,in=120, looseness=1] node[above, yshift=-1mm, color=black]{} (sent2-2-7);
			\draw[chainLine, ->] (sent2-2-3) to [out=60,in=120, looseness=1] node[above, yshift=-1mm, color=black]{} (sent2-2-8);

			\node[wordnode, above= of sent2-1-4, yshift=10mm] (root1) {\Large \texttt{root}};
			\draw[chainLine, -] (root1) to  (sent2-1-4);
			
			\node[wordnode, above= of sent2-2-3, yshift=10mm] (root2) {\Large \texttt{root}};
			\draw[chainLine, -] (root2) to  (sent2-2-3);
			
% 			\node[wordnode, above= of sent2-1-1, xshift=5mm, yshift=14mm] (da) {\LARGE\texttt{Dist=7}};
% 			\node[wordnode, above= of sent2-2-1, xshift=5mm, yshift=15mm] (db) {\LARGE\texttt{Dist=3}};
			\node[wordnode, above= of sent2, yshift=16mm] (da) {\Large{Dist(\textit{awesome}, \textit{lunch}) = 4,~~  Dist(\textit{awesome}, \textit{menu}) = 3}};
			\node[wordnode, below= of da, yshift=-36mm] (db) {\Large{Dist(\textit{awesome}, \textit{lunch}) = 2,~~  Dist(\textit{awesome}, \textit{menu}) = 1}};

			\node[wordnode, below=of sent2-1-4](a){\Large (a)};
			\node[wordnode, below=of sent2-2-4](b){\Large (b)};
		\end{tikzpicture} 
		
	}
% 	\vspace{-3mm}%%cr version
	\caption{Two examples from the Twitter and Rest16 dataset to illustrate the difference between a dependency parse tree (a) and an aspect-centric tree (b). Red words indicate the aspect words of the sentence.}
	\label{fig:case}
\end{figure}
\subsection{Case Study}
\label{sec:cs}
To gain further insight on our induced aspect-centric tree, we use Chu-Liu-Edmonds' algorithm~\citep{edmonds1967optimum} to extract the aspect-centric trees, where each tree is expressed by a weighted adjacency matrix as shown in Equation (\ref{equ:rootprob}). 
We selected two examples from the Twitter and Rest16 datasets, whose sentiments can be correctly predicted by our ACLT model. 
Overall we observe that aspect-centric trees differ from the standard dependency trees in the types of dependencies they create which tend to be of shorter length.

Specifically, as shown in Figure~\ref{fig:case} top (a),  the root of the dependency parse tree is the word ``\textit{looks}'' which is inconsistent with the aspect word ``\textit{google}'' or ``\textit{wave}'', and the key opinion word ``\textit{interesting}'' requires three-hop and two-hop interactions to establish a connection with each of the  two aspect words respectively. However, as shown in Figure~\ref{fig:case} top (b), our aspect-centric tree is rooted in the aspect word ``\textit{wave}''\footnote{Here ``wave'' is chosen as the root because it has the highest probability (i.e., $\bm{P}_i^r$ in Equation~\ref{equ:rootprob}).}. In addition, we observe that the opinion words and aspect words can be connected by two-hop and one-hop interactions through our aspect-centric tree,  which is more effective than the number of interaction hops needed in the dependency parse tree. 
% We also have similar observations for the second case, which illustrated in the bottom of the Figure~\ref{fig:case}.

We also have similar observations for the second case. 
Illustrated in Figure~\ref{fig:case} bottom (a),  the distance between the aspect words ``\textit{lunch}" and ``\textit{menu}" and the critical opinion word ``\textit{awesome}" is four-hops and three-hops, respectively, in the parse tree.   
In contrast,  Figure~\ref{fig:case} bottom (b) shows that in the aspect-centric tree extracted by our model, the distances between aspect and opinion words are one-hop and two-hops, which is closer than the distance in the standard dependency parse tree.

\section{Related Work} 

\paragraph{Aspect-based sentiment analysis.}Early efforts on aspect-based sentiment focused on predicting  polarity by employing attention mechanism \citep{bahdanau2014neural} to model interactions between aspect and context words~\citep{wang2016attention,chen2017recurrent,liu2017attention,li2018transformation,wang2018target}. More recently, neural pre-trained language models, for instance,  BERT~\citep{devlin2019bert} enabled ABSA to achieve promising results.  For example, \citet{sun2019utilizing} manually constructed auxiliary sentences using the aspect word to convert ABSA into a sentence-pair classification task. \citet{huang2019syntax} propagated opinion features from syntax neighborhood words to the aspect words, in a BERT-based model. Another line of work in ABSA focused on leveraging the explicit dependency parse trees to model the relationships between context and aspect words.~\citet{zhang2019aspect} and~\citet{sun2019aspect}  used GCNs to integrate dependency tree information to capture structural and contextual information simultaneously for aspect-based sentiment analysis. \citet{wang2020relational} greedily reshaped the dependency parse trees by using manual rules to obtain the task-specific syntactic structures.
% Unlike previous works which rely on an external parser, we focus on treating the dependency parser as a latent variable and induce it in an end-to-end fashion.
% \vspace{-6mm}
\paragraph{Latent variable induction.}
\textcolor{black}{Latent variable models~\citep{maillard2017jointly,kim2017structured,niculae2018towards,mensch2018differentiable,liu2018learning,zou-19-qt} have gained much popularity in building Natural Language Processing (NLP) pipelines and discovering task-specific linguistic structures~\citep{kim2018tutorial,martins2019latent}.}
% to increase their capability of leading better generalization and interpretability~\citep{kim2018tutorial,martins2019latent}. 
The crucial obstacle of designing structured latent variable models is that they typically involve computing an ``argmax'' (i.e., searching the highest-scoring discrete structure, such as a parse tree) in the middle of a computation graph.
% Since this operation has null gradients it can not be back-propagated out of the box for training~\citep{martins2019latent}. 
End-to-end approaches directly replace the ``argmax'' approach by introducing a continuous relaxation for which the exact gradient can be computed and back-propagated normally.
For example,~\citet{nan2020reasoning} and~\citet{guo2020learning} used marginal inference to construct latent structures to improve information aggregation in the relation extraction task.
More in line with our work,~\citet{chen2020inducing} constructed task-specific structures by developing a gate mechanism to dynamically combine the parse tree information and HardKuma structure. 
Our work differs from this prior work in three main aspects. 
First, we construct the aspect-specific tree for inference without relying on an external parser.
% Second, we impose constraints on the aspect words by using token-level signals to explicitly improve the probability of aspect words becoming the root when inducing a matrix tree.
Second, we facilitate the interactions between target and opinion by introducing an explicit supervision to adaptively adjust the aspect to be the root in an end-to-end fashion.
% indicating that the aspect term is identified as the root of the latent tree.
% Third, we compute the multi-root marginal probabilities which enables our model to dynamically build the latent tree based on multiple aspect words, allowing the model to meticulously capture the complex interactions.
% Third, we compute each aspect word's probability to become the root rather than the single of them, which enables our model to lower the boundary of inferencing root for MTT in the training process.
Third, we compute each aspect word's probability to become the root which enables our model to reduce the search space of inferring root for MTT in the training process.
% To our knowledge, we are the first to link up aspects with opinions through the specifically designed latent tree that imposes root constraints. 
% dynamically build the latent tree based on multiple aspect words, allowing the model to meticulously capture the complex interactions.
%%%conclu
\section{Conclusion and Future Work}
% we compute each aspect word's probability to become the root rather than computing the probability of a single aspect word.
% Thus, the roots of the trees induced by MTT could only be the aspect words rather than the others.
% In other words, we lower the boundary of inferencing root for MTT in the training process.
% We introduce a novel aspect-centric latent tree (ACLT) model to establish effective relationships between aspect words and opinion words in an aspect-based sentiment analysis task. 
In this paper, we propose to use {\em Aspect-Centric Latent Trees} (ACLT) which are specifically tailored for the ABSA  task to link up aspects with opinion words in an end-to-end fashion.
Experiments on five benchmark datasets show the effectiveness of our model. The qualitative and quantitative analysis illustrate that our model is able to improve the probability of aspect words becoming the root of the sentence by imposing root constraints.
Moreover, thorough analysis demonstrates our model shortens the average distances between aspect and opinions by at least 19\% on the SemEval Restaurant14 dataset.
To the best of our knowledge, we are the first to link up aspects with opinions through the specifically designed latent tree that imposes root constraints.
One possible future direction is to apply the proposed approach to other sentiment analysis tasks, such as aspect triplet extraction~\cite{xu-etal-2020-position}.
% In the future, we plan to extend our method to other pre-trained language models such as RoBERTa~\cite{liu2019roberta} and GPT-2~\cite{radford2019language}. 
% We also intend to test our model's performance with other tasks, such as summarization and event extraction. 
% To our knowledge, we are the first to provide an effective solution to address the root inconsistency issue by linking up aspects with opinions through our specifically designed latent tree that imposes root constraints. 

% In the future, we intend to test the model's performance with other tasks, such as summarization and event extraction.
% We  intend to test our model's performance with other tasks, such as summarization and event extraction. 
\section*{Acknowledgements}
We thank Xiaochi Wei, Qian Liu and Ankit Garg and the anonymous reviewers for their comments.
This work was supported by the National Key R\&D Program of China under Grant No. 2020AAA0106600, supported by the National Natural Science Foundation of China (Grant No. U19B2020), sponsored by CCF-Tencent Open Research Fund (CCF-Tencent IAGR20200103) and the funding from China Scholarship Council No. 201906030188. 
% and supported by the National Key R\&D Program of China under Grant No. 2020AAA0106600.
This research is also supported by Ministry of Education, Singapore, under its Academic Research Fund (AcRF) Tier 2 Programme (MOE  AcRF  Tier  2 Award No: MOE2017-T2-1-156). Any opinions, findings and conclusions or recommendations expressed in this material are those of the authors and do not reflect
the views of the Ministry
of Education, Singapore.
\bibliography{emnlp21_arxiv}%anthology,
\bibliographystyle{acl_natbib}
\appendix
% \section*{Appendix}
% \renewcommand\thesection{\Alph{section}}
\section{Statistics of root inconsistency}
\label{app:A}
We count the number of sentences where the aspect word is inconsistent with the roots of its three different structures for all five datasets. Table~\ref{tab:root inconsistent} shows the details.
% Table generated by Excel2LaTeX from sheet 'Analysis'
% \begin{table}[htbp]
%   \centering
% %  \setlength{\tabcolsep}{3.9mm}{
%  \resizebox{1\linewidth}{!}{
%     \begin{tabular}{lrrrrr}
%     \toprule
%           Trees & \multicolumn{1}{l}{Lap14} & \multicolumn{1}{l}{Rest14} & \multicolumn{1}{l}{Rest15} & \multicolumn{1}{l}{Rest16} & \multicolumn{1}{l}{Twitter} \\
%     \midrule
%     Parser & 591   & 1077  & 484   & 551   & 631 \\
%     MTT   & 542   & 1018  & 463   & 533   & 514 \\
%     ACLT  & 213   & 882   & 202   & 195   & 353 \\
%     Total & 638   & 1120  & 542   & 616   & 692 \\
%     \bottomrule
%     \end{tabular}}%
%       \caption{Statistics of sentences where the aspect word is inconsistent with the roots of its three different structures.}
%   \label{tab:root inconsistent}%
  
% \end{table}%
% Table generated by Excel2LaTeX from sheet 'Analysis'
\begin{table}[htbp]
  \centering
  \resizebox{1\linewidth}{!}{
    \begin{tabular}{lrrrrr}
    \toprule
    Tree  & \multicolumn{1}{c}{Lap14} & \multicolumn{1}{c}{Rest14} & \multicolumn{1}{c}{Rest15} & \multicolumn{1}{c}{Rest16} & \multicolumn{1}{c}{Twitter} \\
    \midrule
    Parser & 591 (93\%)   & 1,077 (96\%)  & 484 (89\%)   & 551 (89\%)   & 631 (91\%) \\
    MTT   & 542 (85\%)   & 1,018 (91\%)  & 463 (85\%)   & 533 (87\%)   & 514 (74\%) \\
    ACLT  & \textbf{213} (\textbf{33\%})   & \textbf{882} (\textbf{79\%})   & \textbf{202} (\textbf{37\%})   & \textbf{195} (\textbf{32\%})   & \textbf{353} (\textbf{51\%}) \\
    \midrule
    Total & 638 \textcolor{white}{(93\%)} & 1,120 \textcolor{white}{(93\%)}   & 542 \textcolor{white}{(93\%)}   & 616 \textcolor{white}{(93\%)}   & 692 \textcolor{white}{(93\%)} \\
    \bottomrule
    \end{tabular}}%
        \caption{Statistics of sentences where the aspect word is inconsistent with the roots of its three different structures.}
  \label{tab:root inconsistent}%
\end{table}%

\section{Hyper-parameters of ACLT}
\label{app:B}
All hyper-parameters are tuned based on the development set. The important hyper-parameters are listed in Table~\ref{tab:hyper}. We employed the uncased version of the BERT model in  PyTorch. 
Following previous conventions, we repeat each experiment three times and average the results, reporting accuracy (Acc.) and macro-f1 ($F_1$).

\begin{table}[htbp!]
  \centering
  \setlength{\tabcolsep}{6.6mm}{
  \resizebox{0.9\linewidth}{!}{

    \begin{tabular}{lr}
    \toprule
    Batch size & 64 \\
    Learning rate & 5.00E-05 \\
    Optimizer & Adam \\
    Max Sequence Length & 96 \\
    Hidden Size & 798 \\
    Hidden Layer & 12 \\
    Dropout probability & 0.1 \\
    \bottomrule
    \end{tabular}}
    }%
      \caption{Hyper-parameters of ACLT.}
  \label{tab:hyper}%
\end{table}%

\end{document}